\documentclass[runningheads]{llncs}

\usepackage{eccv}

\usepackage{eccvabbrv}

\usepackage{graphicx}
\usepackage{booktabs}
\usepackage{multirow}
\usepackage[table,xcdraw]{xcolor}
\usepackage[normalem]{ulem}
\useunder{\uline}{\ul}{}

\usepackage[accsupp]{axessibility}  

\usepackage{hyperref}

\usepackage{orcidlink}

\begin{document}

\title{MultiEmo-Bench: Multi-label Visual Emotion Analysis for Multi-modal Large Language Models} 

\author{Tianwei Chen\inst{1}\orcidlink{0000-0002-2544-7744} \and
Takuya Furusawa\inst{1} \and
Yuki Hirakawa\inst{1} \and
Ryotaro Shimizu\inst{1}\orcidlink{0000-0002-4841-1824} \and
Mo Fan\inst{1}\orcidlink{0009-0008-0681-3391} \and
Takashi Wada\inst{1}
}

\authorrunning{T. Chen et al.}

\institute{ZOZO NEXT Inc.}

\maketitle

\begin{abstract}
  This paper introduces a multi-label visual emotion analysis benchmark dataset for comprehensively evaluating the ability of multimodal large language models (MLLMs) to predict the emotions evoked by images. 
  Recent user studies report an unintuitive finding: humans may prefer the predictions of MLLMs over the labels in existing datasets. We argue that this phenomenon stems from the suboptimal annotation scheme used in existing datasets, where each annotator is shown a single candidate emotion for each image and judges whether it is evoked or not. This approach is clearly limited because a single image can evoke multiple emotions with varying intensities. As a result, evaluations based on these datasets may underestimate the capabilities of MLLMs, yet an appropriate benchmark for evaluating such models remains lacking. To address this issue, we introduce a new multi-label benchmark dataset for visual emotion analysis toward MLLMs evaluation. We hire $20$ annotators per image and ask them to select all emotions they feel from an image. Then, we aggregate the votes across all annotators, providing a more reliable and representative dataset labeled with a distribution of emotions. The resulting dataset contains $10,344$ images with $236,998$ valid votes across eight emotions. Based on this benchmark dataset, we evaluate several recent models, including Qwen3-VL, OpenAI's GPT, Gemini, and Claude. We assess model performance on both dominant emotion prediction and emotion distribution prediction.
  Our results demonstrate the progress achieved by recent MLLMs while also indicating that substantial room for improvement remains. Furthermore, our experiments with LLM-as-a-judge show that the method does not consistently improve MLLMs' performance, indicating its limitations for the subjective task of visual emotion analysis.
  Our dataset and evaluation code will be released in~\url{https://github.com/Tianwei3989/MultiEmo-Bench}.
\end{abstract}

\section{Introduction}
\label{sec:intro}

\begin{figure}[tb]
  \centering
  \includegraphics[width=12cm]{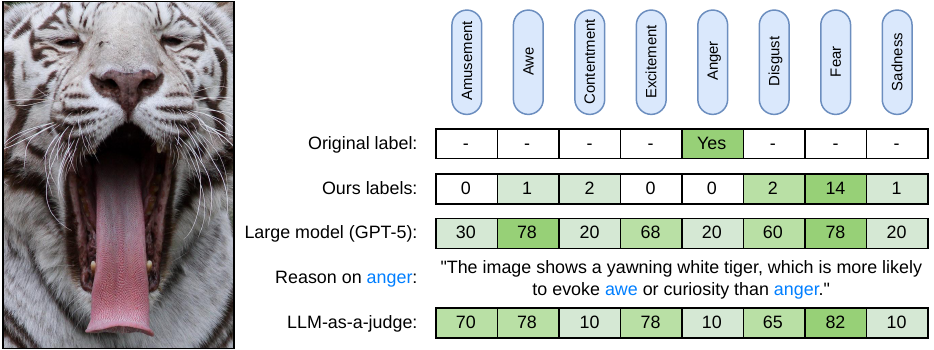}
  \caption{
  We annotate a visual emotion analysis benchmark dataset across all candidate emotions and reveal the inaccurate labels from the original dataset. This dataset is used to evaluate MLLMs and the LLM-as-a-judge method in visual emotion analysis.
  }
  \label{fig:intro}
\end{figure}

Multi-modal large models, such as OpenAI's GPT~\cite{gpt4o,gpt5,gpt51}, Gemini~\cite{gemini25,gemini3}, and Claude~\cite{claude4}, have demonstrated strong progress in understanding humans, sparking increasing interest in exploring how these models can understand emotional concepts~\cite{EmoViT,EmoArt,EEmo,EmoVerse,BhattacharyyaW25,WhyWeFeel,AffectGPT,OVMER,HumanAesExpert,EmotiCrafter}.
Among these evaluations, the visual emotion analysis measures the model’s ability to predict the emotions evoked by an image, a fundamental yet challenging task given the subjective nature of emotional perception~\cite{mikels,two_decade}.

In recent years, evaluations of visual emotion analysis have primarily been conducted in a classification setting, where models are asked to predict a single emotion for each image. However, a recent study~\cite{BhattacharyyaW25} finds that the ground-truth labels in prior datasets~\cite{abstract,FI,EmoSet} can be noisy. User studies reported in this work show that human participants may prefer predictions from large models over the dataset ground-truth labels, suggesting that the original annotations may be inaccurate or incomplete.
We argue that this phenomenon is caused by the verification-based annotation method used in previous datasets, where annotators are given one candidate emotion for each image and verify whether they can feel it or not. This method is not ideal for images that can evoke multiple emotions; \eg,  \cref{fig:intro} shows that the original label for the provided image is \textit{anger}, but we find that some people can feel other emotions such as \textit{awe}, \textit{contentment}, \textit{disgust}, \textit{fear}, and \textit{sadness}.
Since the model evaluations~\cite{FI,EmoSet,EmoViT} are the multi-class classification (\ie, choose one from all candidate emotions), the models' performance would be underestimated if they predict the dominant emotions rather than the labeled emotions.

One solution to this problem is to evaluate models on datasets that are annotated with multiple emotions per image. While several attempts have been made to construct such resources, we find that they are either limited in scale~\cite{abstract,emotion6} or their emotion distributions are highly imbalanced, often skewed toward a single emotion~\cite{LDLs}. Therefore, a large-scale and balanced benchmark is necessary for the rigorous evaluation of MLLMs.

Based on this observation, we propose a new visual emotion analysis benchmark by relabeling existing single-label datasets based on our multi-label annotation scheme. We allow 20 annotators per image to vote for any emotion(s) they perceive from an image, and aggregate them across all annotators to form emotion distributions. The full dataset consists of $10,344$ images, matching the scale of existing datasets~\cite{LDLs} while offering richer and more balanced label distributions.

By this benchmark dataset, we conduct a comprehensive evaluation on recent popular MLLMs, including OpenAI's GPT~\cite{gpt4o,gpt5,gpt51}, Gemini~\cite{gemini25,gemini3}, Claude~\cite{claude4}, and Qwen3-VL~\cite{qwen3vl}. We explore the models' capacity for visual emotion analysis in two tasks: a dominant emotion classification and an emotion distribution prediction.
Among these evaluations, we obtain multiple notable insights, including the identification of underestimation caused by labels from the previous dataset and the observation that older models (\eg, GPT-4o) may outperform newer models (\eg, GPT-5.1).
Furthermore, we conduct a verification study on LLM-as-a-judge to explore whether it can further improve the models' performance.
Our results show that the LLM-as-a-judge does not achieve clear improvement, 
and the model's performance decreases on several metrics, indicating that this strategy may not be consistently helpful for visual emotion analysis.

Overall, the contributions of this paper can be summarized as follows:
\begin{itemize}
  \item We introduce a multi-label visual emotion analysis benchmark that maintains a similar data scale to previous datasets while providing richer annotations and a more balanced label distribution.
  \item We conduct comprehensive evaluations of MLLMs on visual emotion analysis, covering both dominant emotion classification and emotion distribution prediction. Our evaluation includes recent popular models, such as OpenAI's GPT~\cite{gpt4o,gpt5,gpt51}, Gemini~\cite{gemini25,gemini3}, Claude~\cite{claude4}, and Qwen3-VL~\cite{qwen3vl}. The results provide several insights, including the underestimation of model performance caused by previous datasets and the observation that newer models do not necessarily outperform older ones.
  \item We further investigate whether the LLM-as-a-judge strategy can improve model performance. Our results show no clear improvement, suggesting that this strategy may offer limited benefits for the subjective task of visual emotion analysis.
\end{itemize}

\section{Related work}

\subsection{Verification-based visual emotion analysis datasets}
Labeling emotion tags to images~\cite{abstract,FI,EmoSet} is a challenging task, as people may feel different emotions from the same image.
Pioneering studies, such as IASPa~\cite{mikels} and ArtPhoto~\cite{abstract}, carefully collect data from reliable sources (\eg, the International Affective Picture System~\cite{ISPA}) and directly use the emotion tags provided by these sources, resulting in relatively small datasets for evaluation.
Based on this, recent studies, such as FI~\cite{FI} and EmoSet~\cite{EmoSet}, apply annotation to expand the scale of the dataset. 
However, during their annotation, the annotators are asked to verify whether they feel a given emotion in the image, rather than to query which emotions they feel across all emotions. 
Such verification-based annotation may result in inaccurate labeling, as there are images that may evoke multiple emotions. 
As shown in \cref{fig:intro}, we note that the original labels from FI~\cite{FI} and EmoSet~\cite{EmoSet} only label one of the possible emotions, and the labeled emotion is not the dominant emotion. 
During evaluation, labels from these datasets may lead to model underestimation if the model selects the dominant emotion while the label shows another.
As a comparison, our dataset is annotated by query across all emotion candidates, resulting in a more appropriate label for model evaluation.

\begin{table}[tb]
\centering
\caption{Comparison in multi-label visual emotion analysis datasets.
}
\label{tab:datasets_comparison}
\renewcommand{\arraystretch}{1.2}
\setlength{\tabcolsep}{9pt} 
\resizebox{0.98\linewidth}{!}{
\begin{tabular}{crrrrrrc}
\toprule
\multicolumn{1}{l}{} & \multicolumn{1}{c}{images}    & \multicolumn{1}{c}{annotator} & \multicolumn{2}{c}{most frequent}                             & \multicolumn{2}{c}{least frequent}                         &                                 \\
\multicolumn{1}{l}{} & \multicolumn{1}{c}{(in total)}  & \multicolumn{1}{c}{(per image)} & \multicolumn{2}{c}{dominant emotion}                          & \multicolumn{2}{c}{dominant emotion}                       & \multirow{-2}{*}{emotion model} \\ \midrule
Abstract~\cite{abstract}             & \cellcolor[HTML]{FCE5CD}280   & 15.00                         & 78                           & (28\%)                         & \cellcolor[HTML]{FCE5CD}9  & \cellcolor[HTML]{FCE5CD}(3\%) & Mikels (8)                      \\
Emotion6~\cite{emotion6}             & \cellcolor[HTML]{FCE5CD}1,980 & 15.00                         & 642                          & (32\%)                         & \cellcolor[HTML]{FCE5CD}31 & \cellcolor[HTML]{FCE5CD}(2\%) & Ekman (6)                       \\
ArtEmis~\cite{artemis1}              & 81,446                        & \cellcolor[HTML]{FCE5CD}5.68  & -                            & -                              & -                          & -                             & Mikels (8)                      \\
ArtEmis 2~\cite{artemis2}            & 52,933                        & \cellcolor[HTML]{FCE5CD}4.92  & -                            & -                              & -                          & -                             & Mikels (8)                      \\
TwitterLDL~\cite{LDLs}           & 10,045                        & 8.00                          & \cellcolor[HTML]{FCE5CD}7,564 & \cellcolor[HTML]{FCE5CD}(75\%) & 248                        & (2\%)                         & Mikels (8)                      \\
FlickrLDL~\cite{LDLs}            & 11,150                        & 11.00                         & \cellcolor[HTML]{FCE5CD}6,633 & \cellcolor[HTML]{FCE5CD}(59\%) & 230                        & (2\%)                         & Mikels (8)                      \\
Ours                 & 10,344                        & 20.00                         & 2,042                         & (20\%)                         & 611                        & (6\%)                         & Mikels (8)                      \\ \bottomrule
\end{tabular}
}
\end{table}

\subsection{Multi-label visual emotion analysis dataset}
Some prior studies~\cite{abstract,emotion6,artemis1,artemis2,LDLs} also note that a single image can evoke multiple emotions.
However, these datasets are not ideal for a comprehensive evaluation of MLLMs.
As shown in Tab.~\ref{tab:datasets_comparison}, we note that Abstract~\cite{abstract} and Emotion6~\cite{emotion6} are relatively small ($280$ images and $1,980$ images, respectively) for the model evaluation.
ArtEmis~\cite{artemis1} and ArtEmis2~\cite{artemis2} contain large-scale image datasets, but only approximately $5$ annotators are hired per image. 
The datasets most closely related to ours are FlickrLDL and TwitterLDL~\cite{LDLs}. However, we observe a strong imbalance in the dominant emotion labels: $59\%$ of images in FlickrLDL and $75\%$ of images in TwitterLDL share the same dominant emotion. As a result, a trivial strategy that always predicts this emotion can achieve high accuracy on both datasets. 
Motivated by these observations, we aim to construct a multi-label visual emotion analysis dataset with a comparable scale while maintaining a more balanced emotion distribution than FlickrLDL and TwitterLDL~\cite{LDLs}.

\subsection{Affective computing evaluations on MLLMs}
Much effort has been made in exploring how MLLMs can understand human emotions, including multimodal emotion recognition~\cite{FindingEmo,WhyWeFeel,AffectGPT,OVMER,Emotion-LLaMA}, image aesthetics analysis~\cite{ShareGPT4V,UniQA,AesBench,AesExpert,HumanAesExpert}, and visual emotion analysis~\cite{EmoArt,EmoVerse,EEmo,BhattacharyyaW25,Socratis}.
Among these studies, Bhattacharyya \etal~\cite{BhattacharyyaW25} study the most similar topic to our work. In that paper, Bhattacharyya \etal explore MLLMs' capacity for visual emotion analysis from several viewpoints, including the accuracy of evoked emotion recognition, model robustness, and typical error types.
However, since most of the evaluations are based on previous verification-based datasets, the conclusions may warrant further scrutiny. 
In the last part of their user study, it appears that GPT-4o~\cite{gpt4o}'s ``wrong'' predictions are preferred more often than the ground truth labels, and further investigation reveals incorrectness among the labels.
Based on this observation, we are motivated to create a dataset with more reliable labels than the previous datasets, to ensure the reliability of evaluation on visual emotion analysis. 
To the best of our knowledge, this paper is the first comprehensive exploration of evaluating MLLMs in multi-label visual emotion analysis.

\section{MultiEmo-Bench}

\begin{figure}[tb]
  \centering
  \includegraphics[width=12cm]{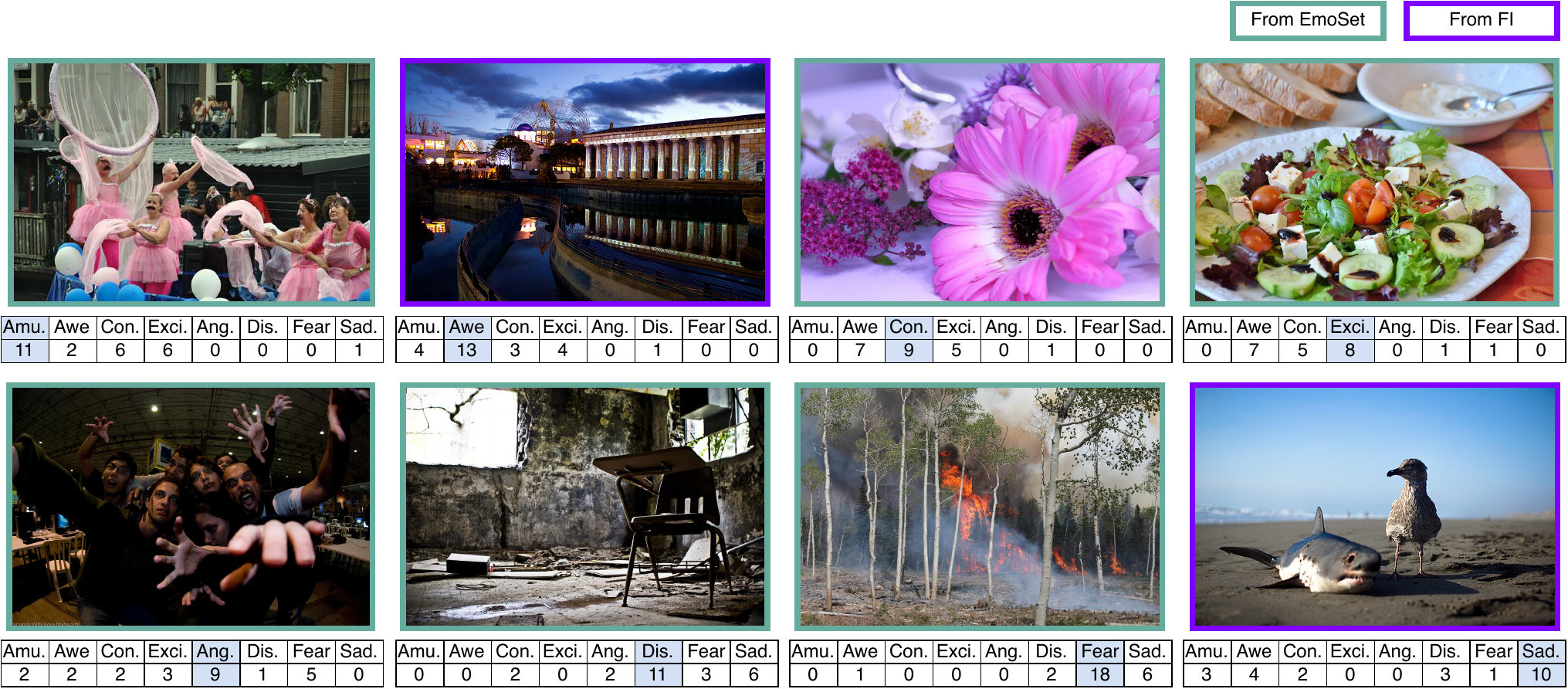}
  \caption{Examples of our dataset. We collect images from EmoSet~\cite{EmoSet} and FI~\cite{FI}. 
  Each image is annotated by $20$ annotators, and each annotator can vote for any of eight Mikels' emotions~\cite{mikels}. 
  The emotions in blue are the dominant emotion.
  }
  \label{fig:dataset}
\end{figure}

MultiEmo-Bench is a benchmark dataset for evaluating visual emotion analysis. This dataset overcomes the weakness of previous verification-based datasets~\cite{FI,EmoSet} while providing richer and more balanced samples per emotion than previous distribution-based datasets~\cite{abstract,emotion6,LDLs}. 
Some examples of our dataset in~\cref{fig:dataset}.
As we introduced in~\cref{sec:intro}, our annotation covers all eight emotion categories, as annotators can vote for any emotion if they feel it. We aggregate all votes per image into an emotion distribution, and the most-voted emotion is taken as the image's dominant emotion.
In this section, we introduce the annotation details, reliability verifications, and the analysis of our dataset.

\subsection{Data selection}
We collect images from the test sets of EmoSet~\cite{EmoSet} ($17,716$ images) and FI~\cite{FI} ($3,474$~\footnote{We randomly sample $15\%$ of the images as the test set, following the setting of FI~\cite{FI}.} images), as they are the largest open-source datasets available to us.
For evaluation efficiency, we focus on images where MLLMs disagree with the labels, as such discrepancies may arise from either incorrect labels or model limitations. To identify these cases, we query three MLLMs—GPT-5~\cite{gpt5}, Gemini-2.5-flash~\cite{gemini25}, and Claude Sonnet 4~\cite{claude4}—on whether the labeled emotion can be evoked by the given image.
During this process, we observe that some images in FI~\cite{FI} are invalid, and some model requests are rejected due to policy restrictions. These invalid or rejected samples are removed before obtaining the final set.
In total, we collect $5,279$ images for which at least one of the models disagrees with the original label. To further examine cases where models may still fail despite agreement, we additionally include $5,065$ images for which all three models agree that the original label is correct (\ie, the complement of the previous set). The resulting dataset therefore contains $10,344$ images in total, corresponding to $48.82\%$ of the images from the combined source datasets.

\subsection{Annotation}
We use Amazon Mechanical Turk (MTurk) to hire annotators for our annotation task.
To each image, we ask each annotator to vote on any emotions they feel from the Mikels'~\cite{mikels} eight emotions, which are \textit{amusement}, \textit{anger}, \textit{awe}, \textit{contentment}, \textit{disgust}, \textit{excitement}, \textit{fear}, and \textit{sadness}. We also provide an option of \textit{none of the above} if the annotator does not feel any of the above eight emotions. 
We do not limit the annotator to vote only one emotion, as we believe that one image may evoke multiple emotions.
For each image, we collect annotations from $20$ annotators, which is the largest number compared to the related work, to enhance both the representativeness and reliability. 
Annotators cannot see any hints, such as the original emotion label, during the annotation.
We emphasize that the annotators should vote for the evoked emotions (\ie, what they feel) instead of the expressed emotion (\ie, what the person in the image feels).

\subsection{Quality control}
\label{sec:quality_control}
To ensure the reliability of our dataset, we apply mainly three verifications before, during, and after the annotation: Before the annotation, we launch qualification tasks to hire qualified annotators and verify the appropriate number of annotators to ensure the representativeness of results. During annotation, we use attention-check questions to ensure annotators remain focused. After the annotation, we conduct an A/B test comparing our labels with those from the original dataset. 

Specifically, regarding the verification of the number of annotators, we collect $193$ annotations per image and compare its distribution with a small subset of annotations. We observe that annotations with $20$ annotators achieve a Pearson’s linear correlation coefficient (PLCC) of 0.836 with 193 annotators’ outputs, suggesting $20$ is an appropriate number of annotators per image. 

In our A/B test, we sample $120$ images where the original labels are different from our dominant emotions. For each image, we ask $50$ participants in MTurk to judge which emotion they feel more from the image, with the options of ``both'' and ``neither''. 
We observe that $56.57\%$ of submissions prefer our dominant emotions over the original labels, while $29.53\%$ of submissions prefer the original labels, indicating that our dominant emotions are also robust to strict testing and preferred over the original labels. There are $8.03\%$ submissions of ``both'' and $5.87\%$ submissions of ``neither'' besides the above two options.
For more details, please refer to the section on quality control in our supplementary materials.

\begin{table}[tb]
\centering
\caption{Comparison in the dominant emotion distribution. The results in green are the most frequent dominant emotions, while those in purple are the least frequent. The summation of images per emotion is slightly larger than the total number of images, as some images have multiple dominant emotions.
}
\label{tab:dominant_emotion_distribution}
\renewcommand{\arraystretch}{1.2}
\resizebox{0.98\linewidth}{!}{
\begin{tabular}{llrrrrrrrrrrrrrrrrlr}
\toprule
{\color[HTML]{333333} }           & {\color[HTML]{333333} } & \multicolumn{16}{c}{{\color[HTML]{333333} images (per emotion)}}                                                                                                                                                                                                                                                                                                                                                                                                                                                                                                                                                                                                                                           & {\color[HTML]{333333} } & \multicolumn{1}{c}{{\color[HTML]{333333} images}}     \\ \cline{3-18}
{\color[HTML]{333333} }           & {\color[HTML]{333333} } & \multicolumn{2}{c}{{\color[HTML]{333333} amusement}}        & \multicolumn{2}{c}{{\color[HTML]{333333} awe}}                                                              & \multicolumn{2}{c}{{\color[HTML]{333333} contentment}}                                                      & \multicolumn{2}{c}{{\color[HTML]{333333} excitement}}       & \multicolumn{2}{c}{{\color[HTML]{333333} anger}}                                                          & \multicolumn{2}{c}{{\color[HTML]{333333} disgust}}                                                        & \multicolumn{2}{c}{{\color[HTML]{333333} fear}}             & \multicolumn{2}{c}{{\color[HTML]{333333} sadness}}          & {\color[HTML]{333333} } & \multicolumn{1}{c}{{\color[HTML]{333333} (in total)}} \\ \cline{3-18} \cline{20-20} 
{\color[HTML]{333333} Abstract~\cite{abstract}}   & {\color[HTML]{333333} } & {\color[HTML]{333333} 33}   & {\color[HTML]{333333} (12\%)} & {\color[HTML]{333333} 29}                           & {\color[HTML]{333333} (10\%)}                         & \cellcolor[HTML]{D9EAD3}{\color[HTML]{333333} 78}   & \cellcolor[HTML]{D9EAD3}{\color[HTML]{333333} (28\%)} & {\color[HTML]{333333} 55}   & {\color[HTML]{333333} (20\%)} & \cellcolor[HTML]{EAD1DC}{\color[HTML]{333333} 9}   & \cellcolor[HTML]{EAD1DC}{\color[HTML]{333333} (3\%)} & {\color[HTML]{333333} 37}                          & {\color[HTML]{333333} (13\%)}                        & {\color[HTML]{333333} 53}   & {\color[HTML]{333333} (19\%)} & {\color[HTML]{333333} 49}   & {\color[HTML]{333333} (18\%)} & {\color[HTML]{333333} } & {\color[HTML]{333333} 280}                            \\
{\color[HTML]{333333} TwitterLDL~\cite{LDLs}} & {\color[HTML]{333333} } & {\color[HTML]{333333} 923}  & {\color[HTML]{333333} (9\%)}  & {\color[HTML]{333333} 265}                          & {\color[HTML]{333333} (3\%)}                          & \cellcolor[HTML]{D9EAD3}{\color[HTML]{333333} 7564} & \cellcolor[HTML]{D9EAD3}{\color[HTML]{333333} (75\%)} & {\color[HTML]{333333} 977}  & {\color[HTML]{333333} (10\%)} & {\color[HTML]{333333} 296}                         & {\color[HTML]{333333} (3\%)}                         & \cellcolor[HTML]{EAD1DC}{\color[HTML]{333333} 248} & \cellcolor[HTML]{EAD1DC}{\color[HTML]{333333} (2\%)} & {\color[HTML]{333333} 348}  & {\color[HTML]{333333} (3\%)}  & {\color[HTML]{333333} 350}  & {\color[HTML]{333333} (3\%)}  & {\color[HTML]{333333} } & {\color[HTML]{333333} 10045}                          \\
{\color[HTML]{333333} FlickrLDL~\cite{LDLs}}  & {\color[HTML]{333333} } & {\color[HTML]{333333} 1147} & {\color[HTML]{333333} (10\%)} & {\color[HTML]{333333} 1406}                         & {\color[HTML]{333333} (13\%)}                         & \cellcolor[HTML]{D9EAD3}{\color[HTML]{333333} 6633} & \cellcolor[HTML]{D9EAD3}{\color[HTML]{333333} (59\%)} & {\color[HTML]{333333} 646}  & {\color[HTML]{333333} (6\%)}  & \cellcolor[HTML]{EAD1DC}{\color[HTML]{333333} 230} & \cellcolor[HTML]{EAD1DC}{\color[HTML]{333333} (2\%)} & {\color[HTML]{333333} 533}                         & {\color[HTML]{333333} (5\%)}                         & {\color[HTML]{333333} 769}  & {\color[HTML]{333333} (7\%)}  & {\color[HTML]{333333} 909}  & {\color[HTML]{333333} (8\%)}  & {\color[HTML]{333333} } & {\color[HTML]{333333} 11150}                          \\
{\color[HTML]{333333} Ours}       & {\color[HTML]{333333} } & {\color[HTML]{333333} 638}  & {\color[HTML]{333333} (6\%)}  & \cellcolor[HTML]{D9EAD3}{\color[HTML]{333333} 2042} & \cellcolor[HTML]{D9EAD3}{\color[HTML]{333333} (20\%)} & {\color[HTML]{333333} 1880}                         & {\color[HTML]{333333} (18\%)}                         & {\color[HTML]{333333} 1942} & {\color[HTML]{333333} (19\%)} & \cellcolor[HTML]{EAD1DC}{\color[HTML]{333333} 611} & \cellcolor[HTML]{EAD1DC}{\color[HTML]{333333} (6\%)} & {\color[HTML]{333333} 1103}                        & {\color[HTML]{333333} (11\%)}                        & {\color[HTML]{333333} 2039} & {\color[HTML]{333333} (20\%)} & {\color[HTML]{333333} 1156} & {\color[HTML]{333333} (11\%)} & {\color[HTML]{333333} } & {\color[HTML]{333333} 10344}                          \\ \bottomrule
\end{tabular}
}
\end{table}

\begin{figure}[tb]
  \centering
  \begin{subfigure}{0.465\linewidth}
    \includegraphics[width=\linewidth]{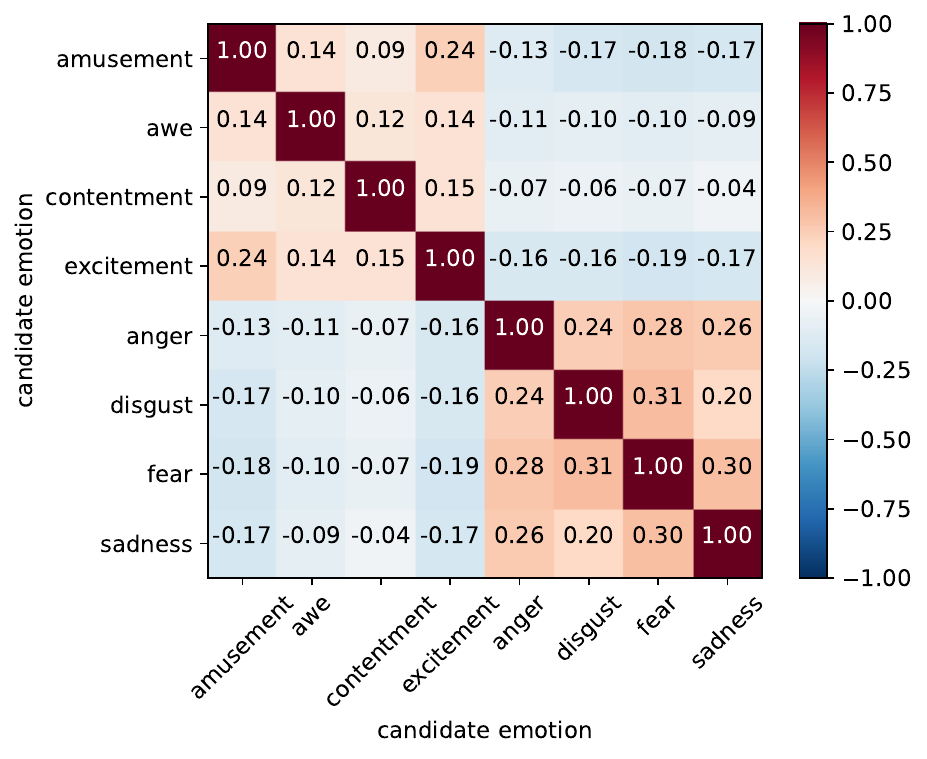}
    \caption{Co-occurrence (NPMI) among emotions.}
    \label{fig:npmi}
  \end{subfigure}
  \hfill
  \begin{subfigure}{0.455\linewidth}
    \includegraphics[width=\linewidth]{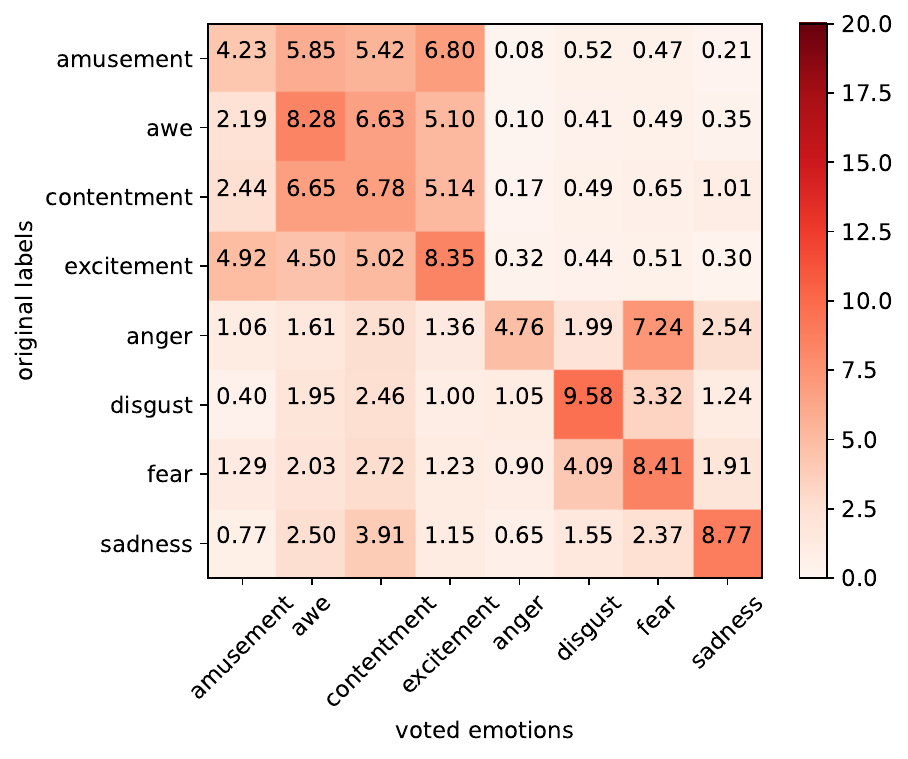}
    \caption{Average votes on each original label.}
    \label{fig:average_voting}
  \end{subfigure}
  \caption{Dataset analysis on annotated emotion co-occurrence and the difference from the original labels. 
  We observe that negative emotions are more likely to co-occur within a single image, and that images originally labeled as \textit{amusement} and \textit{anger} are frequently voted as other emotions by annotators.}
  \label{fig:data_analysis}
\end{figure}

\subsection{Dataset analysis}
\subsubsection{Basic statistic} 
In general, we collect $20$ annotations for each of $10,344$ images, resulting in $206,880$ annotations in total. 
Since the annotators are allowed to vote multiple emotions for each image, we finally collect $236,998$ votes, with an average of $22.91$ per image. 
There are $962$ ($9.3\%$) images with multiple dominant emotions.
Among our $10,344$ images, $10,278$ ($99.36\%$) images have at least one dominant emotion that is voted by at least $5$ annotators.
However, only $4,907$ ($47.44\%$) images have more than $10$ votes for their dominant emotion, indicating the subjectivity of visual emotion analysis.

We apply split-half reliability analyses with Spearman's rank correlation coefficient (SRCC) and Pearson's linear correlation coefficient (PLCC) to assess the annotator agreement in our dataset.
The dataset achieves PLCC of $0.716$ and SRCC of $0.666$, indicating a moderately strong positive linear relationship between the two groups of annotators and suggesting that our annotations show relatively strong agreement among annotators.

\subsubsection{Emotion distribution} 
We explore the distribution of images across the dominant emotions.
As shown in \cref{tab:dominant_emotion_distribution}, the largest emotion category accounts for only about $20\%$ of the data, which is relatively small compared to the previous datasets~\cite{abstract,LDLs}.
Besides, we have more images on the emotion categories of \textit{awe}, \textit{excitement}, \textit{anger}, \textit{disgust}, \textit{fear}, and \textit{sadness}.

\subsubsection{Emotion co-occurrence}
To further explore how often emotion pairs co-occur, we calculate the normalized point-wise mutual information (NPMI) across all images to evaluate their co-occurrence.
Intuitively, the higher (up to $1$) the NPMI score between two emotions, the more often these two emotions are voted, and vice versa. The minimum NPMI score is $-1$, and the NPMI score around $0$ indicates the neutral co-occurrence (\ie, two emotions could be voted together by chance).
As shown in \cref{fig:npmi}, we observe that the emotions among both the positive emotion group and the negative emotion group show a positive co-occurrence with each other. 
The emotions in the negative group have higher co-occurrence scores, indicating that these emotions are more likely to co-occur in a single image.
The co-occurrence between positive and negative emotions ranges from $-0.07$ to $-0.19$, suggesting that positive emotions rarely co-occur with negative emotions, which is intuitive.

\subsubsection{Difference to the original labels}
We observe that only $48.49\%$ of the dominant emotion in our annotation is the same as the original labels. 
To further analyze the difference, we calculate how much the annotators vote differently from the original labels, as shown in \cref{fig:average_voting}.
The verification is made by calculating the average votes on each original label. For example, the $5.85$ in the top-left of \cref{fig:average_voting} suggests an average vote of $5.85$ on \textit{awe} when the original label is \textit{amusement}.
Given a original label $e_i$ and voted emotion $e_j$, the average vote $V_{e_i, e_j}$ can be calculated by:
\begin{equation}
    V_{e_i, e_j} = \frac{1}{|\mathcal{I}_{e_i}|}\sum_{x \in \mathcal{I}_{e_i}}\sum_{a=1}^{|\mathcal A|}\mathbf{1}(l_{x,a}=e_j),
\end{equation}
where $\mathcal{I}_{e_i}$ denotes the set of images with the original label $e_i$, and $l_{x,a}$ represents the vote from annotator $a$ for image $x$ in our collected annotations.
This analysis allows us to assess how our annotations differ from the original labels. 
According to \cref{fig:average_voting}, the images with the original label of \textit{contentment}, \textit{excitement}, \textit{disgust}, \textit{fear}, and \textit{sadness} are highly voted on the original labels, which indicates that these labels may receive more agreements. 
Images originally labeled as \textit{amusement} tend to receive votes for other positive emotions (\eg, \textit{excitement}) as well, and the images originally labeled as \textit{anger} tend to receive votes for \textit{fear}.
This phenomenon may be related to noisy labels in the original datasets, as some labels are more about the \textit{expressed} emotion than the \textit{evoked} emotion. The same observation is also introduced in the recent study~\cite{BhattacharyyaW25}.

\section{Evaluation for multi-modal large language models}
\label{sec:evaluation}
In this section, we introduce our explorations on how MLLMs perform in visual emotion analysis, which mainly includes three parts:
First, we explore the models' performance in terms of the dominant emotion, which is the same evaluation that most current visual emotion analysis studies~\cite{FI,EmoSet,EmoVerse} use.
Second, we evaluate the models' performance by the distribution similarity on all emotions, aiming for a more comprehensive evaluation that includes the non-dominant emotions. 
Third, we evaluate how LLM-as-a-judge can improve the models when knowing other models' outputs. We aim to explore the validity and improvement of LLM-as-a-judge for visual emotion analysis, as it is increasingly used to build large-scale benchmarks~\cite{EmoArt,EmoVerse}.

\subsection{Baseline models}
In this experiment, we evaluate $11$ MLLMs that are widely used in recent studies, including the GPT series (GPT-4o~\cite{gpt4o}, GPT-5-nano~\cite{gpt5}, GPT-5~\cite{gpt5}, and GPT-5.1~\cite{gpt51}), Gemini series (Gemini-2.5-flash~\cite{gemini25} and Gemini-3-flash-preview~\cite{gemini3}), Claude Sonnet 4~\cite{claude4}, and Qwen3-VL series~\cite{qwen3vl} (Qwen3-VL-2B, Qwen3-VL-4B, Qwen3-VL-8B, and Qwen3-VL-32B).

\begin{table}[tb]
\centering
\caption{Precision results on dominant emotion. We prepare a single-dominant-emotion version of our labels for a fair comparison with the original labels.}
\label{tab:dominant_results}
\renewcommand{\arraystretch}{1.4}
\resizebox{.99\linewidth}{!}{ 
\begin{tabular}{lrrrrlrrrrlrrlr}
\hline
                   & \multicolumn{4}{c}{Qwen3-VL}                                                                                                          & \multicolumn{1}{c}{} & \multicolumn{4}{c}{GPT}                                                                                                                        &  & \multicolumn{2}{c}{Gemini}                                            &  & \multicolumn{1}{c}{Claude}      \\ \cline{2-5} \cline{7-10} \cline{12-13} \cline{15-15} 
                   & \multicolumn{1}{c}{2B}          & \multicolumn{1}{c}{4B}          & \multicolumn{1}{c}{8B}          & \multicolumn{1}{c}{32B}         &                      & \multicolumn{1}{c}{4o}          & \multicolumn{1}{c}{5-nano}      & \multicolumn{1}{c}{5}                    & \multicolumn{1}{c}{5.1}         &  & \multicolumn{1}{c}{2.5-flash}   & \multicolumn{1}{c}{3-flash-preview} &  & \multicolumn{1}{c}{Sonnet 4}    \\ \cline{2-5} \cline{7-10} \cline{12-13} \cline{15-15} 
Original label     & 25.78\%                         & 43.12\%                         & 41.78\%                         & 40.65\%                         &                      & 40.38\%                         & 28.32\%                         & \textbf{44.54\%}                         & 41.10\%                         &  & 39.51\%                         & 40.06\%                             &  & 30.40\%                         \\
Our label (single) & \cellcolor[HTML]{FCE5CD}24.38\% & \cellcolor[HTML]{FCE5CD}41.97\% & \cellcolor[HTML]{FCE5CD}41.14\% & \cellcolor[HTML]{D9EAD3}41.48\% &                      & \cellcolor[HTML]{D9EAD3}42.99\% & \cellcolor[HTML]{D9EAD3}29.16\% & \cellcolor[HTML]{D9EAD3}\textbf{47.91\%} & \cellcolor[HTML]{D9EAD3}45.19\% &  & \cellcolor[HTML]{D9EAD3}40.76\% & \cellcolor[HTML]{D9EAD3}40.41\%     &  & \cellcolor[HTML]{D9EAD3}32.41\% \\
Our label          & \cellcolor[HTML]{B6D7A8}26.51\% & \cellcolor[HTML]{B6D7A8}44.90\% & \cellcolor[HTML]{B6D7A8}44.07\% & \cellcolor[HTML]{B6D7A8}44.35\% &                      & \cellcolor[HTML]{B6D7A8}45.90\% & \cellcolor[HTML]{B6D7A8}31.62\% & \cellcolor[HTML]{B6D7A8}\textbf{51.05\%} & \cellcolor[HTML]{B6D7A8}48.14\% &  & \cellcolor[HTML]{B6D7A8}43.62\% & \cellcolor[HTML]{B6D7A8}43.20\%     &  & \cellcolor[HTML]{B6D7A8}35.10\% \\ \hline
\end{tabular}
}
\end{table}

\subsection{Evaluation on the dominant emotion}
\label{sec:exp_dominant}

In this experiment, we evaluate the model's performance by its precision on three different label sets: the original labels, our labels, and a single-label version of our labels. 
Since our labels contain images with multiple dominant emotions, we prepare a single-label version of our labels by randomly selecting only one of the dominant emotions as the valid label to ensure a fair comparison with the original labels.
We ask models to assign a score (0–100) on “How much people will feel the given emotion” for each of the eight emotions, as well as one sentence of reason as a verification of the output. The emotion with the largest score is considered the predicted dominant emotion. 
Specifically, when the model predicts multiple dominant emotions, each correct prediction receives only fractional credit (\eg, $1/k$ when $k$ emotions are predicted). 
We query for each emotion rather than querying models on all eight emotions at once, since some models may have position bias in the exhibit order of emotions.
The calculation of the precision can be represented as:
\begin{equation}
    \mathrm{Prec}=\frac{1}{|\mathcal{I}|}
    \sum_{x \in \mathcal{I}}
    \frac{|\arg\max_{e_i \in \mathcal{E}}s_x(e_i)\;\cap\;
    \arg\max_{e_i \in \mathcal{E}} n_x(e_i)|}{\left|\arg\max_{e_i \in \mathcal{E}} s_x(e_i)\right|},
\end{equation}
where $s_x(e_i)$ is the predicted score assigned by the model to emotion label $e_i$ for image $x$, and $n_x(e_i)$ denotes the number of annotator votes for label $e_i$ on image $x$. 
The outputs of both $\arg\max_{e_i \in \mathcal{E}}s_x(e_i)$ and $\arg\max_{e_i \in \mathcal{E}} n_x(e_i)$ are sets of candidate emotions.
The results are shown in~\cref{tab:dominant_results}.
According to~\cref{tab:dominant_results}, we have the following observations:
\begin{itemize}
  \item \emph{Original labels tend to cause an underestimation of models’ performance.} Comparing results between our full labels and the original labels, we observe that all models receive higher accuracy with our labels than with the original labels. Models (\eg, GPT-5 and GPT-5.1) with higher scores receive even larger gains than those models (\eg, GPT-5-nano) with lower scores. These results suggest an underestimation caused by the original labels, which are likely affected by the incomplete labeling method.
  \item \emph{GPT-5 achieves the highest accuracy, yet the task remains challenging.} GPT-5 performs best at identifying the emotions most commonly perceived by annotators. However, its accuracy reaches only $51.05\%$, indicating that the task remains difficult and that substantial room for improvement remains.
\end{itemize}

\subsection{Evaluation on emotion distribution}
\label{sec:exp_distribution}

We convert each emotion’s votes into an emotion distribution to serve as ground truth for a more comprehensive model performance comparison in this experiment. The evaluation of emotion distributions is less straightforward and interpretable than that of the dominant emotion, but it captures more details about the model’s performance on non-dominant emotions, which are more robust when a single image evokes multiple emotions with different strengths.

We evaluate the same baseline models using the same inference procedure in~\cref{sec:exp_dominant}. Specifically, we establish a baseline of human performance, as humans may have diverse responses due to the subjectivity of emotion. The human baseline is established by comparing the distributions of emotions between two randomly half-separated groups of annotators. We run $10$ random separations and use the average score as the final result.

We prepare eight metrics for evaluating the models' performance, which are Chebyshev distance (Cheb), Canberra distance (Canber), Wasserstein distance (Wasser), Kullback-Leibler divergence (KLdiv), Jensen–Shannon divergence (JSdiv), cosine similarity (Cosine), Spearman's rank correlation coefficient (SRCC), and Pearson's linear correlation coefficient (PLCC).
Since the predictions and the annotation votes are on different numerical scales, we normalize the votes and scores by dividing their potential maximum value (\ie, $20$ for votes and $100$ for scores) before calculating each metric. 
The human baseline on the Kullback-Leibler divergence and the Jensen–Shannon divergence are not applicable, as these metrics are not comparable when the distribution of the referred side (\ie, the $Q$) is different.

\begin{table}[tb]
\centering
\caption{Baseline results on emotion distribution prediction. The best results are \textbf{bold}, while the worst results are {\ul underlined}.
}
\label{tab:baseline_results}
\renewcommand{\arraystretch}{1.3}
\setlength{\tabcolsep}{7pt}   
\resizebox{.99\linewidth}{!}{
\begin{tabular}{lrrrrrrrr}
\hline
                           & \multicolumn{1}{c}{Cheb ↓} & \multicolumn{1}{c}{Canber ↓} & \multicolumn{1}{c}{Wasser ↓} & \multicolumn{1}{c}{KLdiv ↓} & \multicolumn{1}{c}{JSdiv ↓} & \multicolumn{1}{c}{Cosine ↑} & \multicolumn{1}{c}{PLCC ↑} & \multicolumn{1}{c}{SRCC ↑} \\ \hline
Qwen-VL-2B                 & {\ul 0.830}                & 4.924                        & 0.343                        & 2.331                       & {\ul 0.467}                 & {\ul 0.642}                  & {\ul 0.345}                & {\ul 0.318}                \\
Qwen-VL-4B                 & 0.684                      & 5.012               & 0.293                        & 1.485                       & 0.335                       & 0.794                        & 0.593                      & 0.604                      \\
Qwen-VL-8B                 & 0.651                      & 4.120                        & 0.209                        & 0.998                       & 0.338                       & 0.775                        & 0.579                      & 0.587                      \\
Qwen-VL-32B                & 0.616                      & 4.976                        & 0.247                        & 1.388                       & 0.345                       & 0.794                        & 0.592                      & 0.586                      \\ \hline
GPT-4o                     & 0.657                      & 4.859                        & 0.268                        & 1.294                       & \textbf{0.322}              & \textbf{0.813}               & \textbf{0.634}             & \textbf{0.615}             \\
GPT-5-nano                 & 0.635                      & {\ul 5.479}                  & {\ul 0.388}                  & {\ul 2.709}                 & 0.394                       & 0.725                        & 0.423                      & 0.459                      \\
GPT-5                      & \textbf{0.578}             & 4.947                        & 0.240                        & 1.308                       & 0.337                       & 0.811                        & 0.631                      & 0.588                      \\
GPT-5.1                    & 0.595                      & 4.920                        & 0.233                        & 1.200                       & 0.328                       & 0.811                        & 0.629                      & 0.588                      \\ \hline
Gemini-2.5-flash           & 0.662                      & \textbf{3.798}               & 0.166                        & \textbf{0.693}              & 0.370                       & 0.727                        & 0.541                      & 0.578                      \\
Gemini-3-flash-preview     & 0.609                      & 4.443                        & \textbf{0.160}               & 0.765                       & 0.352                       & 0.753                        & 0.559                      & 0.568                      \\ \hline
Claude Sonnet 4            & 0.648                      & 4.989                        & 0.245                        & 1.404                       & 0.358                       & 0.756                        & 0.519                      & 0.556                      \\ \hline
\textit{Human (half-half)} & 0.559                      & 2.647                        & 0.097                        & -                           & -                           & 0.821                        & 0.700                      & 0.656                      \\ \hline
\end{tabular}
}
\end{table}

\begin{figure}[tb]
  \centering
  \includegraphics[width=12cm]{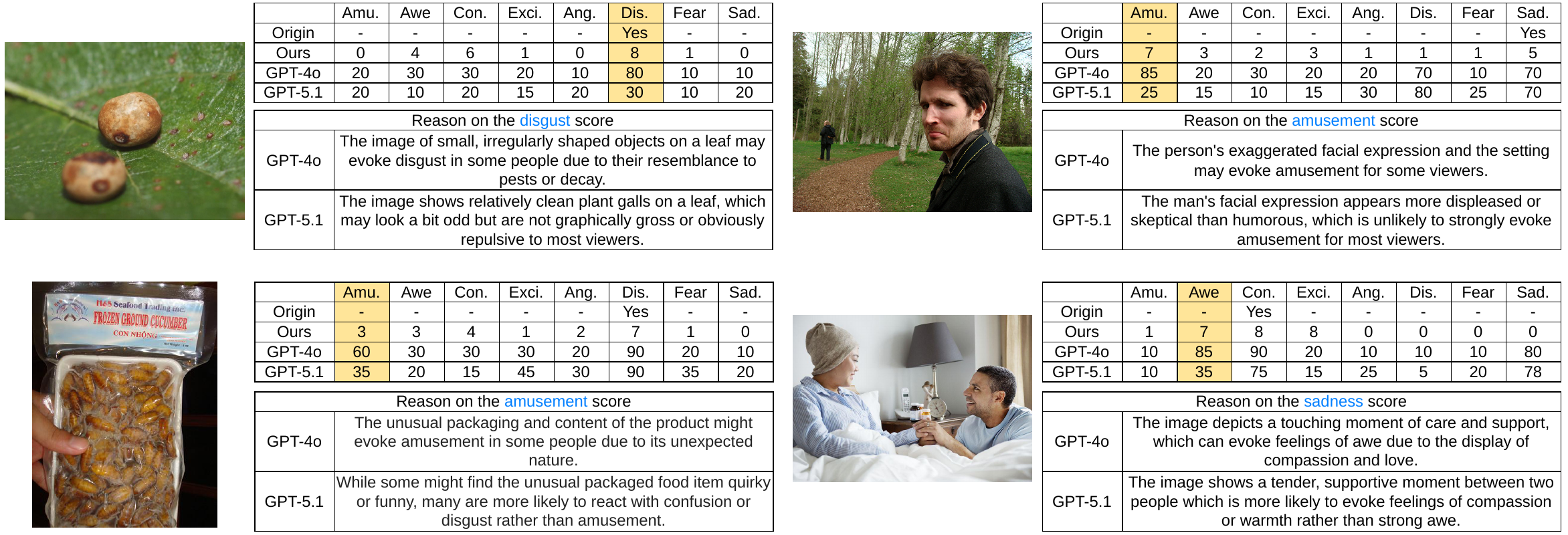}
  \caption{Examples of GPT-4o and GPT-5.1 outputs. The scores in orange are the most different between these two models. We observe that GPT-5.1 suppresses some emotions when other emotions are stronger.}
  \label{fig:example_llm}
\end{figure}

Our results are shown in~\cref{tab:baseline_results}.
By viewing these results, we make the following observations:
\begin{itemize}
  \item \emph{GPT-4o performs the best in most of the metrics.} We observe that GPT-4o gets the best scores among four of eight metrics. Specifically, we find that GPT-4o beats GPT-5.1 across five of eight metrics, despite GPT-5.1 being announced as smarter and featuring emotional features such as ``warmer'' and ``playful''.
  Specifically, GPT-4o outperforms GPT-5.1 in Canber and SRCC, indicating that GPT-4o scores emotions closer to the human annotations and performs better at ranking the strength of eight emotions. However, GPT-4o is defeated by GPT-5.1 in the Cheb metric. Since Chebyshev distance shows the longest distance in all dimensions, this result indicates that GPT-5.1 performs better in the worst prediction, which means GPT-5.1 makes fewer mistakes than GPT-4o.
  \item \emph{The lightest models perform worst in all metrics.} We observe that the lightest models, Qwen3-VL-2B and GPT-5-nano, perform clearly worse than any of the other models in our experiment. 
  By manually checking the outputs, we note that GPT-5-nano tends to predict high scores across all emotions: $7,526$ samples are predicted to evoke more than four emotions, and two samples have eight dominant emotions. We recommend avoiding using the lightest models, such as GPT-5-nano, in the visual emotion analysis task. 
  \item \emph{Larger models may not perform better}. According to the results in the Qwen series, we find that Qwen3-VL-32B is not always better than Qwen3-VL-4B or Qwen3-VL-8B. In the metrics of Canber, JSdiv, Cosine, PLCC, and SRCC, Qwen3-VL-4B performs slightly better than Qwen3-VL-32B. These results may indicate that, when the model is large enough, expanding the model scale may not yield a clear improvement.
  \item \emph{Similar performance in ranking, but different performance in scoring.} 
  We observe that all models except the two lightest models have similar results in SRCC, ranging from $0.556$ to $0.615$, which indicates that they have similar capacity in deciding whether an emotion is evoked more strongly than another. However, the PLCC results are diverse, ranging from $0.519$ to $0.634$, indicating that these models differ in their ability to score the strength of emotions.
  \item \emph{Though progressing, there is still room for models to improve.} We observe that most of the models have achieved a performance that is close to the human baseline. According to PLCC and SRCC, all models except the two lightest models have achieved scores greater than $0.5$, indicating that the predictions have a moderately strong positive correlation to the human annotations.
  However, we also note gaps between models and the human baseline in the Cheb, Canber, and Wasser metrics, indicating room for model improvement.
\end{itemize}

Furthermore, we make a qualitative analysis of GPT-4o and GPT-5.1 to further explore the performance of these two models.
Some examples are shown in~\cref{fig:example_llm}.
We note that GPT-5.1 often assigns low scores to emotions when it considers that another emotion is dominant. Such behavior may cause a focus shift as the model may focus on other emotions instead of judging whether the target emotion is probably evoked by the image. We also observe that GPT-4o captures the rich nuances of emotions, including the stimuli (\eg, pest in the top-left image) and global settings (\eg, the scene in top-right and bottom-right images), which tends to help GPT-4o predict answers aligning with human feelings

\subsection{Evaluation of LLM-as-a-judge}
\label{sec:exp_laaj}

Furthermore, we explore how LLM-as-a-judge can improve performance in multi-label visual emotion analysis.
To ensure a similar knowledge volume, we evaluate the LLM-as-a-judge performance on GPT-5~\cite{gpt5}, Gemini-2.5-flash~\cite{gemini25}, and Claude Sonnet 4~\cite{claude4}, which were all released around May to August in $2025$.
We collect the outputs (\ie, the predicted scores and the reasons) for these models from~\cref{sec:exp_dominant}, and ask each of these three models to provide a final score after concluding all of the corresponding outputs. 
Specifically, to ensure the models think enough before their final prediction, we also ask the models to analyze each output from both views: ``why the output could be correct'' and ``why the output could be wrong''. 
In general, the model thus needs to provide six sentences of analysis (\ie, answers of ``why correct'' and ``why wrong'' for each of the three models), one sentence of conclusion, and one score of how the emotion is likely to be evoked by the given image.
To prevent the model from biasing towards its output, we ensure the model cannot see which model the output comes from.

\begin{table}[tb]
\centering
\caption{LLM-as-a-judge results. We compare the performance between the straight predictions (\ie, w/o LLM-as-a-judge) and the prediction with LLM-as-a-judge to verify the effect of LLM-as-a-judge.
}
\label{tab:laaj_results}
\renewcommand{\arraystretch}{1.3}
\setlength{\tabcolsep}{5pt}
\resizebox{.98\linewidth}{!}{
\begin{tabular}{lllrlrrrrrrrr}
\toprule
                                   &          &  & \multicolumn{1}{c}{Dominant}    &  & \multicolumn{8}{c}{Distribution}                                                                                                                                                                                                                              \\ \cline{4-4} \cline{6-13} 
                                   &          &  & \multicolumn{1}{c}{Acc ↑}       &  & \multicolumn{1}{c}{Cheb ↓}    & \multicolumn{1}{c}{Canber ↓}  & \multicolumn{1}{c}{Wasser ↓}  & \multicolumn{1}{c}{KLdiv ↓}   & \multicolumn{1}{c}{JSdiv ↓}   & \multicolumn{1}{c}{Cosine ↑}  & \multicolumn{1}{c}{PLCC ↑}    & \multicolumn{1}{c}{SRCC ↑}    \\ \cline{4-4} \cline{6-13} 
                                   & straight &  & 51.05\%                         &  & 0.646                         & 3.995                         & 0.145                         & 1.118                         & 0.341                         & 0.800                         & 0.631                         & 0.588                         \\
\multirow{-2}{*}{GPT-5}            & w/ judge &  & \cellcolor[HTML]{FCE5CD}46.97\% &  & \cellcolor[HTML]{FCE5CD}0.686 & \cellcolor[HTML]{FCE5CD}4.142 & \cellcolor[HTML]{D9EAD3}0.132 & \cellcolor[HTML]{D9EAD3}0.949 & \cellcolor[HTML]{FCE5CD}0.356 & \cellcolor[HTML]{FCE5CD}0.766 & \cellcolor[HTML]{FCE5CD}0.587 & \cellcolor[HTML]{FCE5CD}0.565 \\ \cline{1-2} \cline{4-4} \cline{6-13} 
                                   & straight &  & 43.62\%                         &  & 0.729                         & 3.697                         & 0.135                         & 0.787                         & 0.379                         & 0.720                         & 0.541                         & 0.578                         \\
\multirow{-2}{*}{Gemini-2.5-flash} & w/ judge &  & \cellcolor[HTML]{FCE5CD}41.46\% &  & \cellcolor[HTML]{FCE5CD}0.747 & \cellcolor[HTML]{D9EAD3}3.652 & \cellcolor[HTML]{FCE5CD}0.141 & \cellcolor[HTML]{FCE5CD}0.832 & \cellcolor[HTML]{FCE5CD}0.392 & \cellcolor[HTML]{FCE5CD}0.710 & \cellcolor[HTML]{FCE5CD}0.526 & \cellcolor[HTML]{FCE5CD}0.561 \\ \cline{1-2} \cline{4-4} \cline{6-13} 
                                   & straight &  & 35.10\%                         &  & 0.762                         & 3.865                         & 0.165                         & 1.079                         & 0.379                         & 0.718                         & 0.519                         & 0.556                         \\
\multirow{-2}{*}{Claude Sonnet 4}  & w/ judge &  & \cellcolor[HTML]{D9EAD3}38.27\% &  & \cellcolor[HTML]{D9EAD3}0.726 & \cellcolor[HTML]{FCE5CD}3.905 & \cellcolor[HTML]{D9EAD3}0.141 & \cellcolor[HTML]{D9EAD3}0.986 & \cellcolor[HTML]{FCE5CD}0.391 & \cellcolor[HTML]{FCE5CD}0.706 & \cellcolor[HTML]{FCE5CD}0.503 & \cellcolor[HTML]{FCE5CD}0.535 \\ \bottomrule
\end{tabular}
}
\end{table}

We evaluate performance based on both accuracy for the dominant emotions and distribution similarity, as in~\cref {sec:exp_dominant,sec:exp_distribution}, and the results are shown in~\cref{tab:laaj_results}.
By viewing these results, we have the following observations:
\begin{itemize}
  \item \emph{LLM-as-a-judge may not always be helpful.} We observe that LLM-as-a-judge does not provide the desired improvement to the models. Instead, models with LLM-as-a-judge often show worse performance than those without LLM-as-a-judge. A possible reason for this phenomenon is that the models with lower performance introduce noise into models with the best performance, while the best models are not powerful enough to denoise and extract useful information from other models due to the subjective task.
  \item \emph{Weakest model gets best benefits, but the improvements are limited.} Claude Sonnet 4 is the model getting the most benefit. The improvement may be due to Claude Sonnet 4's lowest performance among the three models. 
  We also observe that this model learns most from other models, as its LLM-as-a-judge predictions are more similar to those of GPT-5 and Gemini-2.5-flash than to its own. The PLCC score between its LLM-as-a-judge predictions and GPT-5's and Gemini-2.5-flash's predictions is higher than the score between its predictions and its LLM-as-a-judge predictions.
  However, we note that the improvements are limited as none of the improved scores defeat the best scores among the three models.
\end{itemize}

\section{Limitations}
\subsection{Ethic.}
We note that there are rising concerns about the ethical considerations of emotion recognition. As emotions are subjective and personal, trying to predict them with a machine learning model may be intrusive. 
We agree that emotion prediction could raise privacy issues and potential risks of model abuse.
Being aware of this, we did our best to address these concerns proactively. 
Before our annotation task, we introduced our privacy policy, including their right to decline and withdraw submissions, to the annotators.
In our experiments, we handled data responsibly and ensured that its use aligns with ethical standards.
Besides, we are planning to inform users of the inherent risks associated with our dataset and ensure they utilize it responsibly.
Furthermore, we are prepared to take swift action, including freezing or deleting portions or the entirety of the dataset, if we identify any significant risks associated with its use.

\subsection{Emotion taxonomy and co-occurrence.}
In this paper, we conduct all our annotations and experiments in English, which limits their applicability for evaluation in other cultural systems. For example, the bottom-left image of \cref{fig:example_llm} shows frozen worms labeled ``frozen ground cucumber'', which may hardly elicit \textit{amusement} when people cannot understand the label. 
Furthermore, we note co-occurrence between certain emotions (\eg, \textit{disgust} and \textit{fear} in \cref{fig:npmi}) from our annotation. This phenomenon also appears in the previous datasets~\cite{abstract,emotion6,LDLs,artemis1,artemis2} and MLLMs' outputs, which may be caused by the taxonomy of Mikels' eight emotions~\cite{mikels}. As a matter of fact, the eight emotions are the top-4 positive emotions and top-4 negative emotions picked by a pilot study in Mikels' study~\cite{mikels}.
We consider that there remains room for further improvement in the visual emotion analysis when a better emotion taxonomy comes from the psychological side.

\section{Conclusion}

In this paper, we conduct a comprehensive evaluation of visual emotion analysis for recent large models. For this purpose, we first introduce a novel benchmark dataset that is considered more reliable and has a more balanced distribution than previous datasets. Our evaluation results show the progress made by recent large models, while indicating that the task remains challenging. Furthermore, we show insights into models’ performance in visual emotion analysis, such as the older model (\eg, GPT-4o) may not be worse than the new one(\eg, GPT-5.1), and LLM-as-a-judge may not be helpful in the visual emotion analysis task.


\section*{Acknowledgements}
We thank Takahiro Kawashima for helpful discussions on the human baseline and evaluation metrics.

%
%
\bibliographystyle{splncs04}
\bibliography{main}

\newpage

\appendix

\section{Annotation details}
\begin{figure}[tb]
  \centering
  \includegraphics[width=12cm]{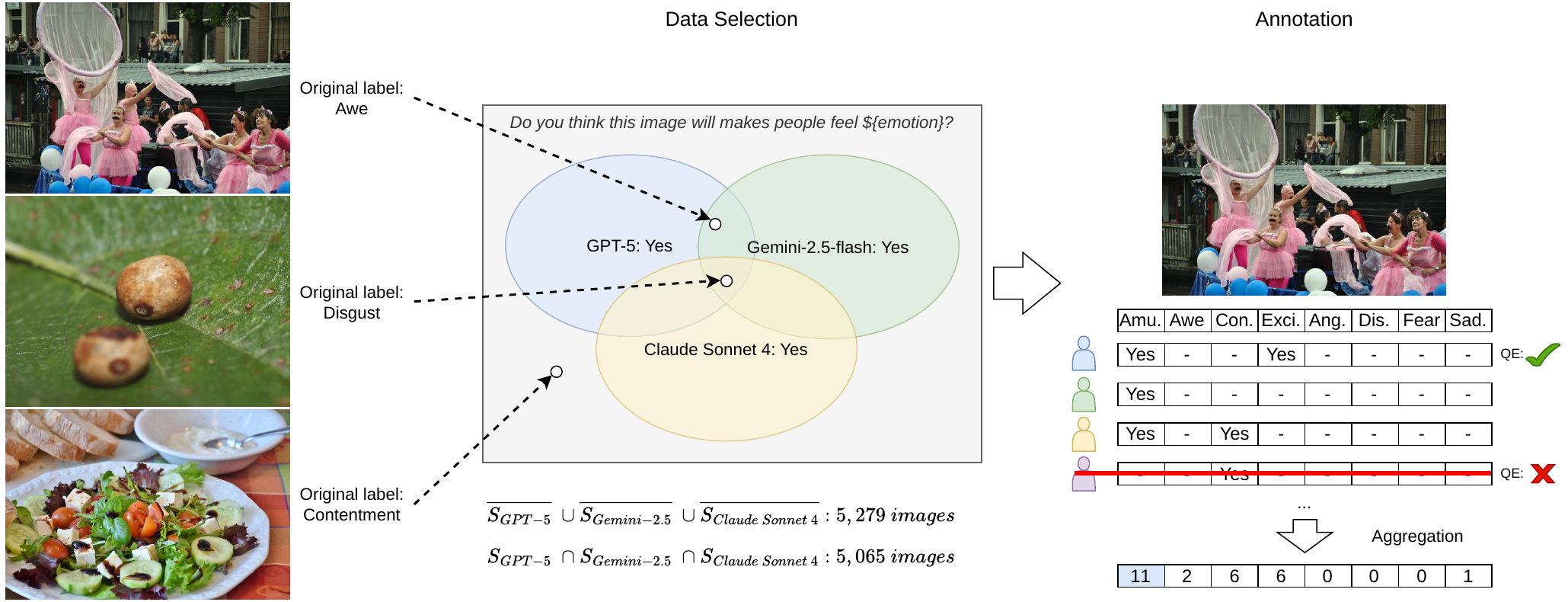}
  \caption{
  Annotation workflow
  }
  \label{fig:workflow}
\end{figure}

The overall annotation workflow is shown in~\cref{fig:workflow}. In this section, we provide further details about our annotation process, including the annotation interface, the quality control procedure, and the A/B test setup.

\begin{figure}[tb]
  \centering
  \includegraphics[width=6cm]{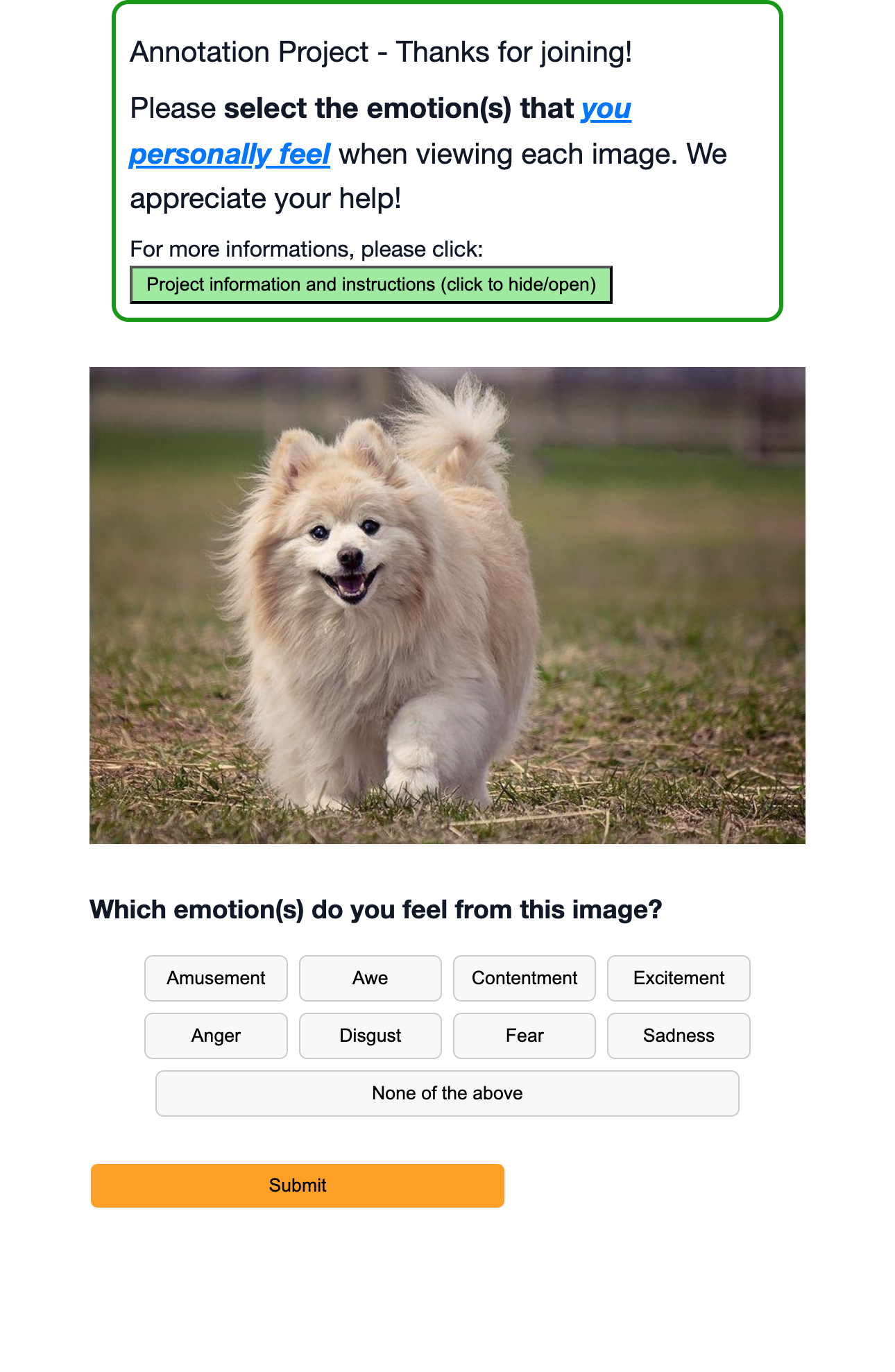}
  \caption{
  Annotation interface
  }
  \label{fig:anno_page}
\end{figure}

\subsection{Annotation interface}

We design a simple HTML interface for the annotation task. As shown in~\cref{fig:anno_page}, annotators can select any emotions they perceive from the image. An additional option, \textit{None of the above}, is provided if annotators do not feel any of the listed emotions.

\begin{figure}[tb]
  \centering
  \begin{subfigure}{0.465\linewidth}
    \includegraphics[width=\linewidth]{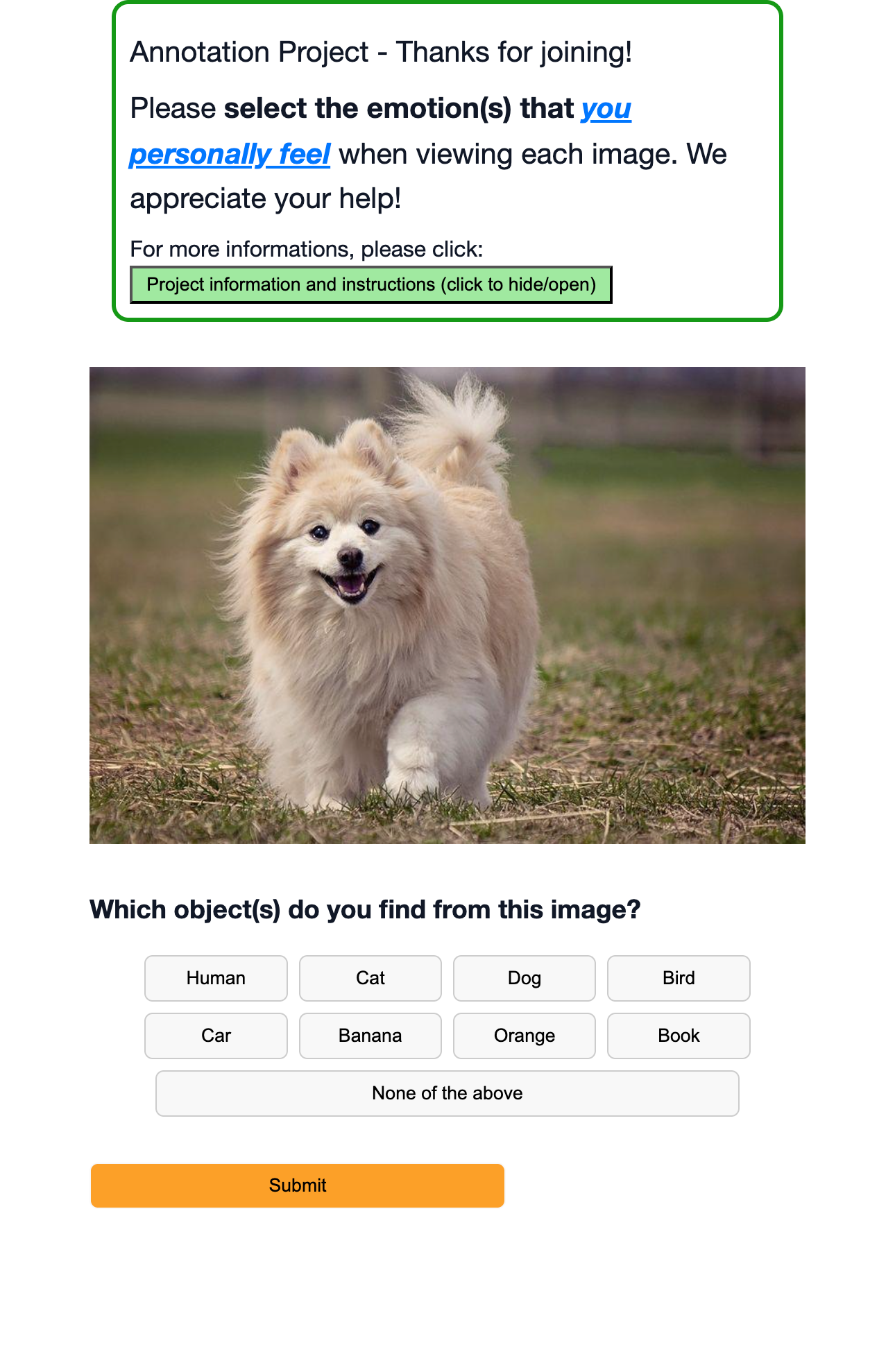}
    \caption{Example of the dummy questions}
    \label{fig:anno_verify_dummy}
  \end{subfigure}
  \hfill
  \begin{subfigure}{0.465\linewidth}
    \includegraphics[width=\linewidth]{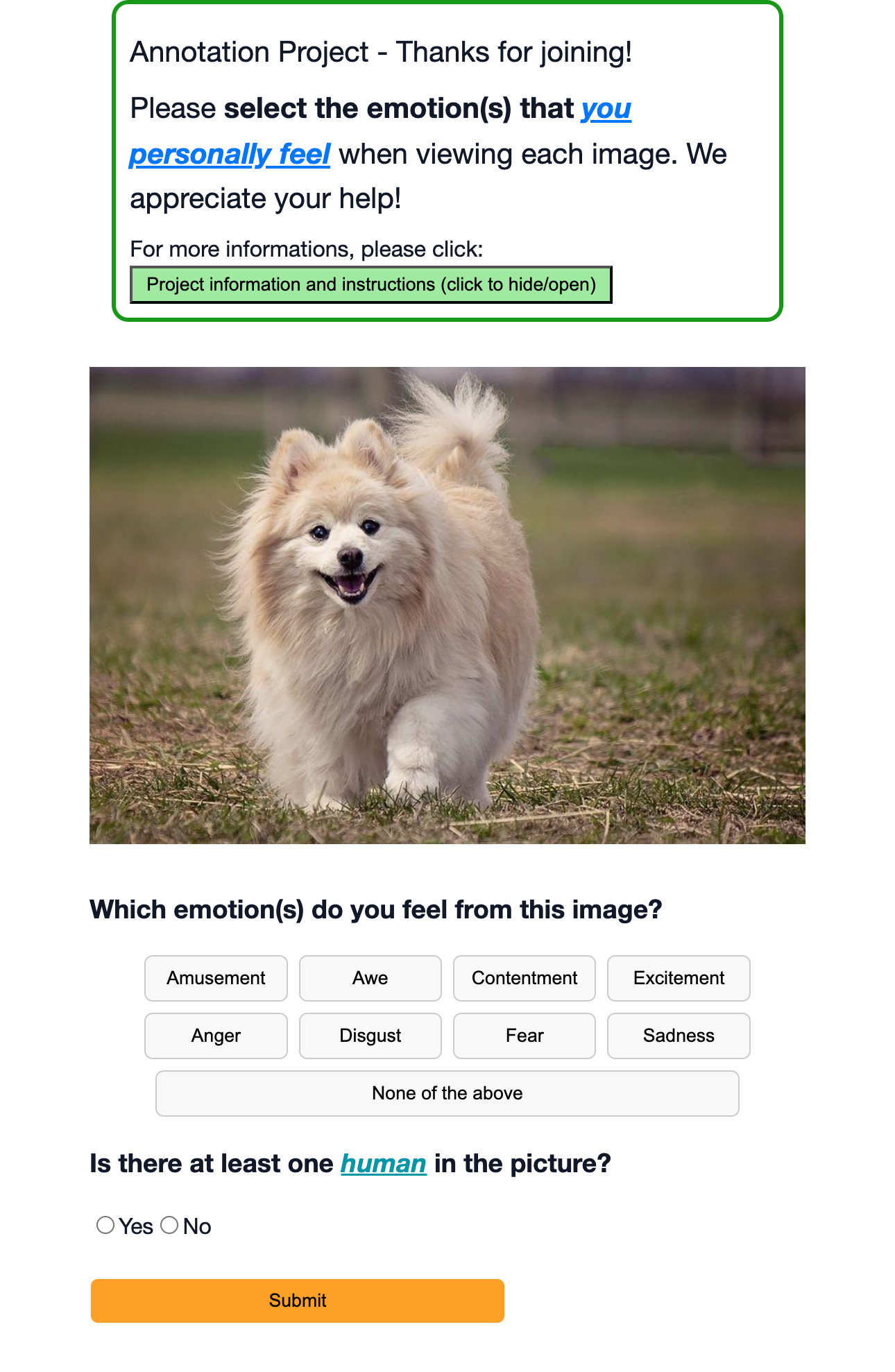}
    \caption{Example of the in-page questions}
    \label{fig:anno_verify_inpage}
  \end{subfigure}
  \caption{Verification questions in the quality control process}
  \label{fig:anno_verify}
\end{figure}

\subsection{Quality control}
\label{sec:quality_control_supp}

We apply quality control both before and during the annotation process to ensure the reliability of the annotations. Before the main annotation begins, we publish a qualification task to recruit annotators. In this task, annotators are asked to select the emotions they perceive from given images, which follows the same procedure as the main annotation task.

As shown in~\cref{fig:anno_verify}, the qualification task includes two types of verification questions to assess annotators' attention and honesty: dummy questions and in-page questions.

For the dummy questions, the annotation task is modified from emotion labeling to object labeling, \ie, from ``which emotion(s) do you feel'' to ``which object(s) do you find.'' Annotators who submit incorrect answers are considered either dishonest or inattentive to the task.

For the in-page questions, annotators are asked an additional question about whether a specific object exists in the same image. This question can be left unanswered when submitting the annotation. Annotators who submit a blank or incorrect answer are considered insufficiently attentive to the task.

We also enforce two MTurk qualification requirements: (1) at least $5,000$ approved Human Intelligence Tasks (HITs), and (2) an approval rate above $95\%$ across all HITs.

During the qualification task, annotators are considered qualified if they meet the following criteria: (1) submit at least $10$ HITs; (2) make no mistakes on either type of verification question (\ie, dummy questions or in-page questions); (3) have fewer than $80\%$ of their submissions assigned to the same emotion; and (4) have fewer than $50\%$ of their submissions labeled as \textit{none of the above}.
To maintain dataset diversity, we divide the $10,344$ images into $11$ batches, and each annotator is allowed to participate in at most $3$ rounds of annotation. As a result, we recruit $442$ qualified annotators from $1,693$ participants.

During the main annotation stage, we include in-page questions in $20\%$ of the annotations and add dummy questions to another $20\%$ of the annotations. Annotators who make errors in more than $5\%$ of their submissions are removed from the dataset.

\subsection{Reliability and representativeness analysis}

To ensure the reliability of our dataset, we apply three analysis steps before, during, and after the annotation process. Before the annotation, we explore the appropriate number of annotators required to obtain representative and reliable results. During the annotation, we assess agreement among annotators. After the annotation, we conduct an A/B test comparing our labels with those from the original dataset.

\begin{figure}[t]
\centering
\begin{minipage}[t]{0.45\linewidth}
    \centering
    \vspace{0pt}
    \includegraphics[width=\linewidth]{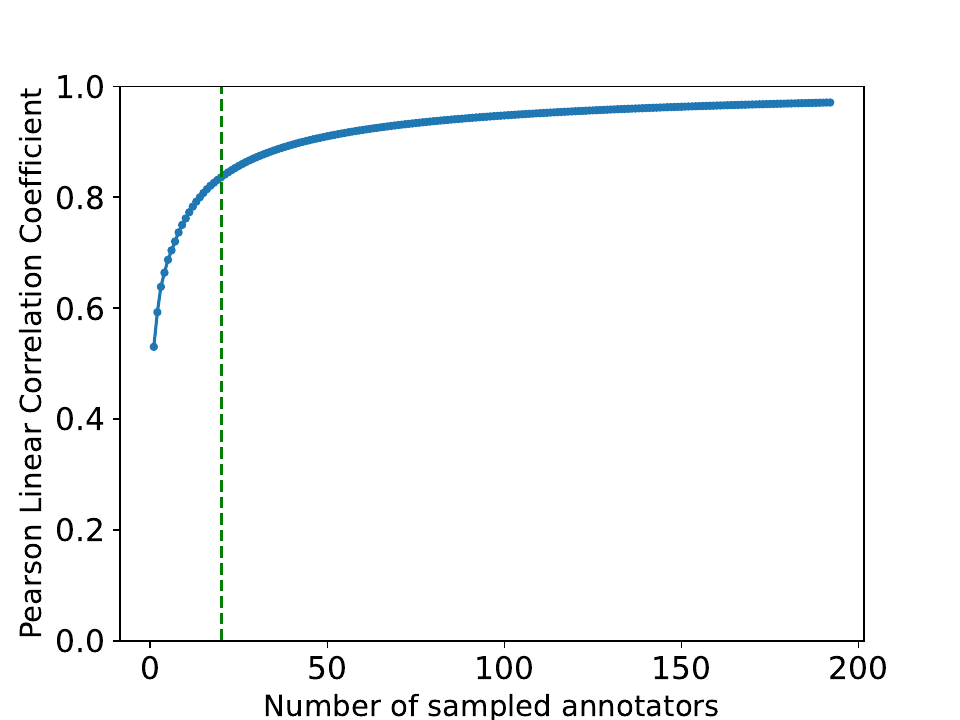}
    \caption{Voting similarity between sampled annotators and all $193$ annotators. The green vertical line shows the number of annotators (\ie, $20$) in our annotation.}
    \label{fig:num_annotator}
\end{minipage}\hfill
\begin{minipage}[t]{0.45\linewidth}
    \centering
    \vspace{0pt}
    \captionof{table}{Results of the A/B test between our dominant emotions and the original labels. The total agreement on our dataset is $64.60\%$, while the total agreement on the original dataset is $37.57\%$.}
    \label{tab:AB_test}
    \begin{tabular}{|cc|cr|}
\hline
\multicolumn{2}{|c|}{\multirow{2}{*}{}}        & \multicolumn{2}{c|}{Origin}                           \\ \cline{3-4} 
\multicolumn{2}{|c|}{}                                 & \multicolumn{1}{c|}{Yes}     & \multicolumn{1}{c|}{No} \\ \hline
\multicolumn{1}{|c|}{\multirow{2}{*}{Ours}} & Yes & \multicolumn{1}{r|}{8.03\%}  & 56.57\%                 \\ \cline{2-4} 
\multicolumn{1}{|c|}{}                           & No  & \multicolumn{1}{r|}{29.53\%} & 5.87\%                  \\ \hline
\end{tabular}
\end{minipage}
\end{figure}

\subsubsection{Number of annotation per image.}
We conduct the same qualification task as introduced in~\cref{sec:quality_control_supp} to explore the number of annotators required to produce a representative annotation. To this end, we publish the qualification task $260$ times to obtain a rich emotion distribution across $90$ images. After removing unqualified annotators and their annotations, the image with the fewest annotations retains $193$ valid annotations.
In this verification experiment, we measure the similarity of emotion distributions $\bar{p}_n$ between an $n$-sample random subset $S_n$ and the full annotation set $\mathcal{A}$ (\ie, the $193$ annotations). Our goal is to identify an appropriate value of $n$ that closely approximates the distribution of $\mathcal{A}$ while remaining small enough to reduce annotation costs.

For each value of $n$, we repeat the sampling process $10$ times with different random subsets $S_n$. We then compute the distribution similarity using Pearson’s linear correlation coefficient (PLCC).
The process could be represented as:
\begin{equation}
    \bar p_n = \frac{1}{T}\sum_{t=1}^T\mathrm{PLCC}(D_{S_{n,t}},D_{\mathcal A}), 
    \quad S_{n,t} \sim\mathrm{Unif}\big(\{S \subset \mathcal{A}, |S|=20 \}\big) 
\end{equation}
, where $T$ denotes the number of repeated trials, which is set to $10$. $D_{S_{n,t}}$ and $D_{\mathcal A}$ denote the aggregated distributions derived from annotators in set $S_t$ and $\mathcal A$, respectively.

As shown in \cref{fig:num_annotator}, annotation with $20$ annotators (\ie, $n = 20$) achieves a PLCC score of $0.836$, and the curve begins to converge, indicating that $n = 20$ is an appropriate number of annotators to ensure representativeness.

\subsubsection{Annotator agreement.}

We apply split-half reliability analysis using Spearman's rank correlation coefficient (SRCC) and Pearson's linear correlation coefficient (PLCC) to assess annotator agreement in our dataset. In this analysis, we split the annotations for each image into two disjoint groups and compute SRCC and PLCC between the two groups.

We repeat the experiment with $10$ different random seeds and report the average scores as the final results. During the annotation process, SRCC ranges from $0.701$ to $0.726$, while PLCC ranges from $0.648$ to $0.665$. In the final dataset, the split-half reliability reaches $0.666$ in SRCC and $0.716$ in PLCC, indicating moderately strong positive correlations between the two groups and suggesting relatively high agreement among annotators given the subjectivity of the task.

\subsubsection{A/B testing.}

After completing the annotation process, we conduct an A/B test to examine whether our annotations are preferred over the original dataset labels. In this test, we compare the dominant emotion from our annotations with the emotion label provided in the original dataset. Annotators are also given the options \textit{both} and \textit{neither} when they feel that both emotions apply or that neither is appropriate.

If our annotations were unreliable, annotators would be expected to prefer either the original label or select \textit{neither}. To ensure a clear comparison, we select images where the dominant emotion in our annotations has at least $7$ more votes than the emotion in the original label. From these candidates, we randomly sample $15$ images for each of the eight emotions, resulting in a test set of $120$ images.

We recruit $50$ annotators per image to ensure the reliability of the test. To prevent answer leakage, we include only qualified annotators who have not previously seen these images and randomize the order of the emotion options during the test.

After the test, we treat the option \textit{both} as ``yes'' for both emotions and the option \textit{neither} as ``no'' for both emotions. The results are reported in~\cref{tab:AB_test}. 

We observe that $64.60\%$ of annotators vote for our dominant emotions, while $37.57\%$ vote for the original labels. These results indicate that our dominant emotion annotations remain robust under strict evaluation and are preferred over the original labels.

\begin{figure}[tb]
  \centering
    \includegraphics[width=0.6\linewidth]{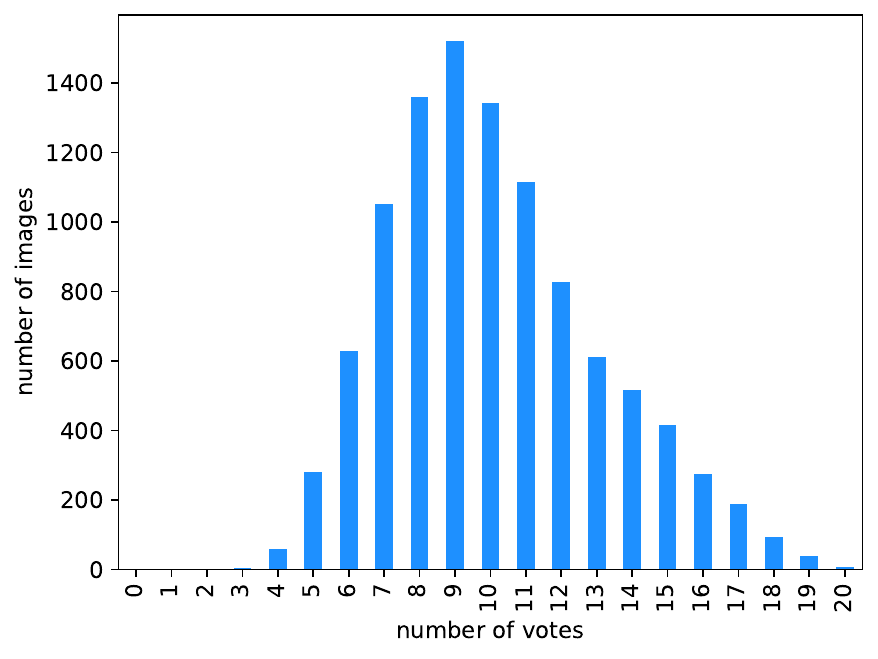}
    \caption{Agreements (votes) on the dominant emotion. $99.36\%$ of the images have at least one dominant emotion that is voted by at least $5$ annotators.}
    \label{fig:major_emo_vote_dist}
\end{figure}

\subsection{Agreements (votes) on the dominant emotions}
We analyze the number of votes received by the dominant emotions, as shown in~\cref{fig:major_emo_vote_dist}.
As reported in the basic statistics section of the data analysis, $10,278$ ($99.36\%$) images have at least one dominant emotion that is supported by at least $5$ annotators. However, only $4,907$ ($47.44\%$) images receive more than $10$ votes for their dominant emotion. This phenomenon may be attributed to the subjective nature of visual emotion analysis, as different individuals may perceive different emotions from the same image.

\section{Experiment details}

\begin{figure}[tb]
  \centering
  \includegraphics[width=12cm]{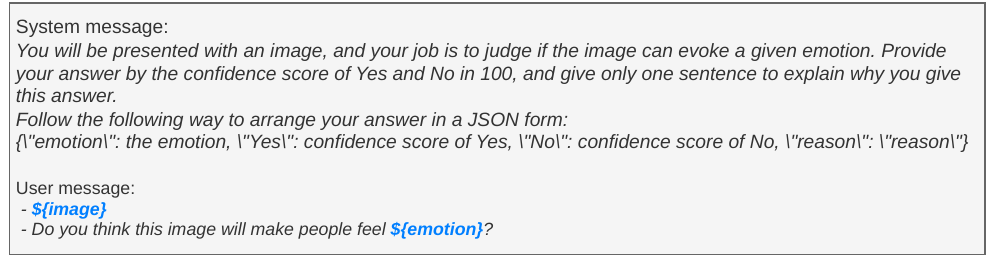}
  \caption{
  Prompt for the straight MLLM inference.
  }
  \label{fig:prompt_infer}
\end{figure}

\begin{figure}[tb]
  \centering
  \includegraphics[width=12cm]{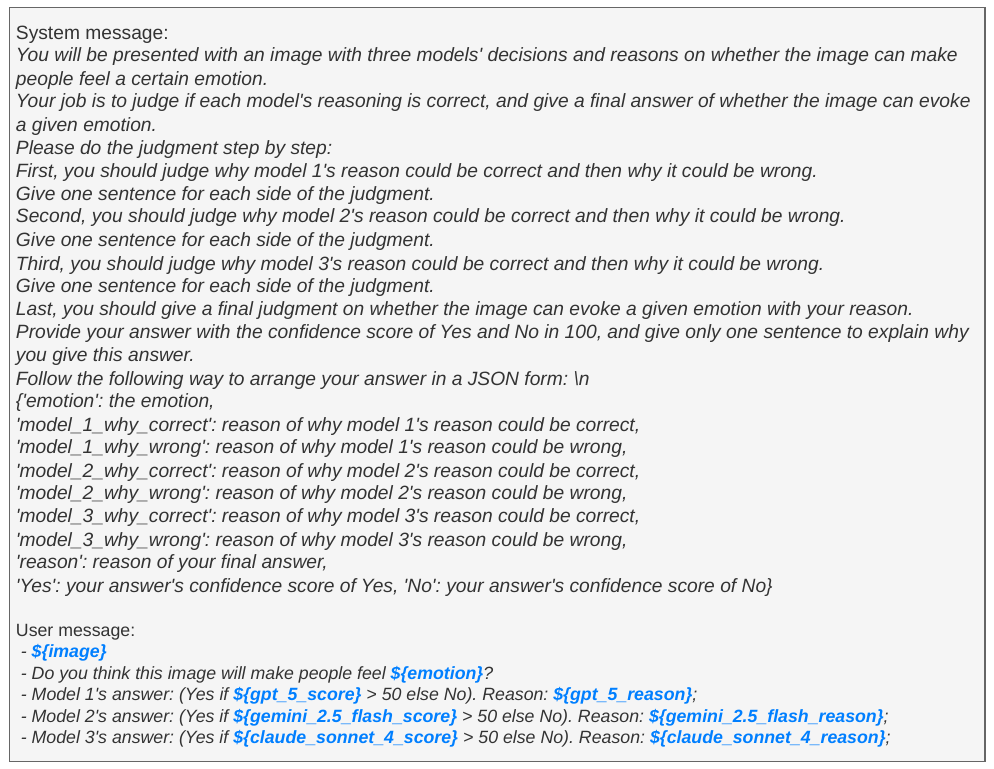}
  \caption{
  Prompt for the LLM-as-a-judge.
  }
  \label{fig:prompt_laaj}
\end{figure}

\subsection{Model versions}
The model versions for each MLLMs are shown follow.

\emph{Qwens:} [Qwen3-VL-2B-Instruct, Qwen3-VL-4B-Instruct, Qwen3-VL-8B-Instruct, Qwen3-VL-32B-Instruct]

\emph{GPTs:} [gpt-4o-2024-11-20, gpt-5-nano-2025-08-07, gpt-5-2025-08-07, gpt-5.1-2025-11-13]

\emph{Geminis:} [gemini-2.0-flash, gemini-3-flash-preview]

\emph{Claude:} [claude-sonnet-4@20250514]

\subsection{Prompts}
We use two types of prompts. The first is designed for direct MLLM inference to estimate how strongly an image evokes a given emotion, as shown in~\cref{fig:prompt_infer}. The second is used for the LLM-as-a-judge setting, where the MLLM first analyzes the outputs from the direct inference before producing the final prediction, as shown in~\cref{fig:prompt_laaj}.

Specifically, we ask ``Do you think this image will make people feel \$emotion?'' rather than ``Do you feel \$emotion from this image?'', as our goal is to evaluate emotions perceived by humans rather than the responses of the MLLMs themselves.

\section{Additional visualization examples}

\begin{figure}[tb]
  \centering
  \includegraphics[width=12cm]{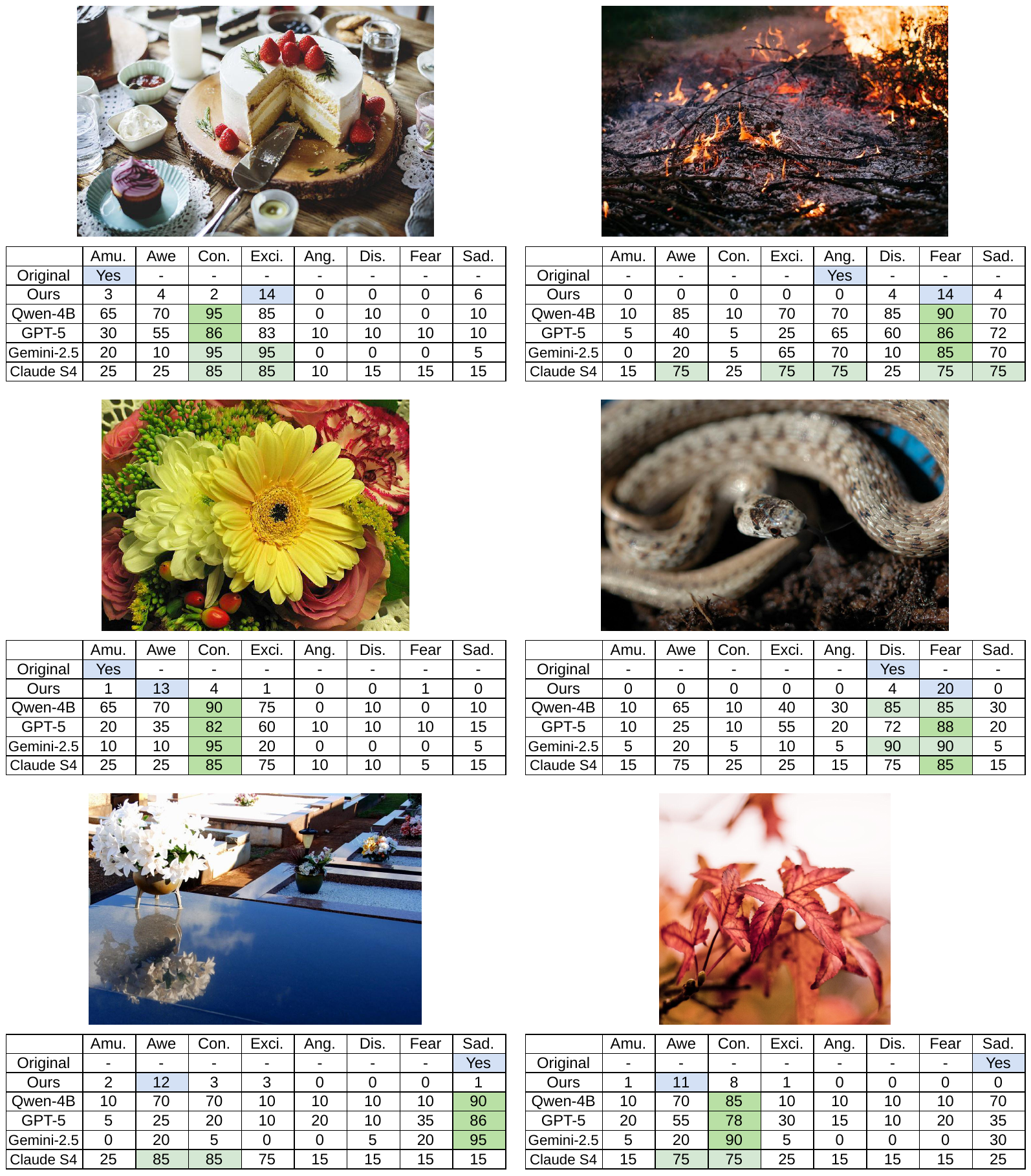}
  \caption{Additional examples of the annotation and the MLLM predictions. The scores in blue are the labeled dominant emotion, while the score in green are the predicted dominant emotions. The light green scores indicate that the model predict multiple dominant emotions on one image.}
  \label{fig:examples_dominant}
\end{figure}

\begin{figure}[tb]
  \centering
  \includegraphics[width=12cm]{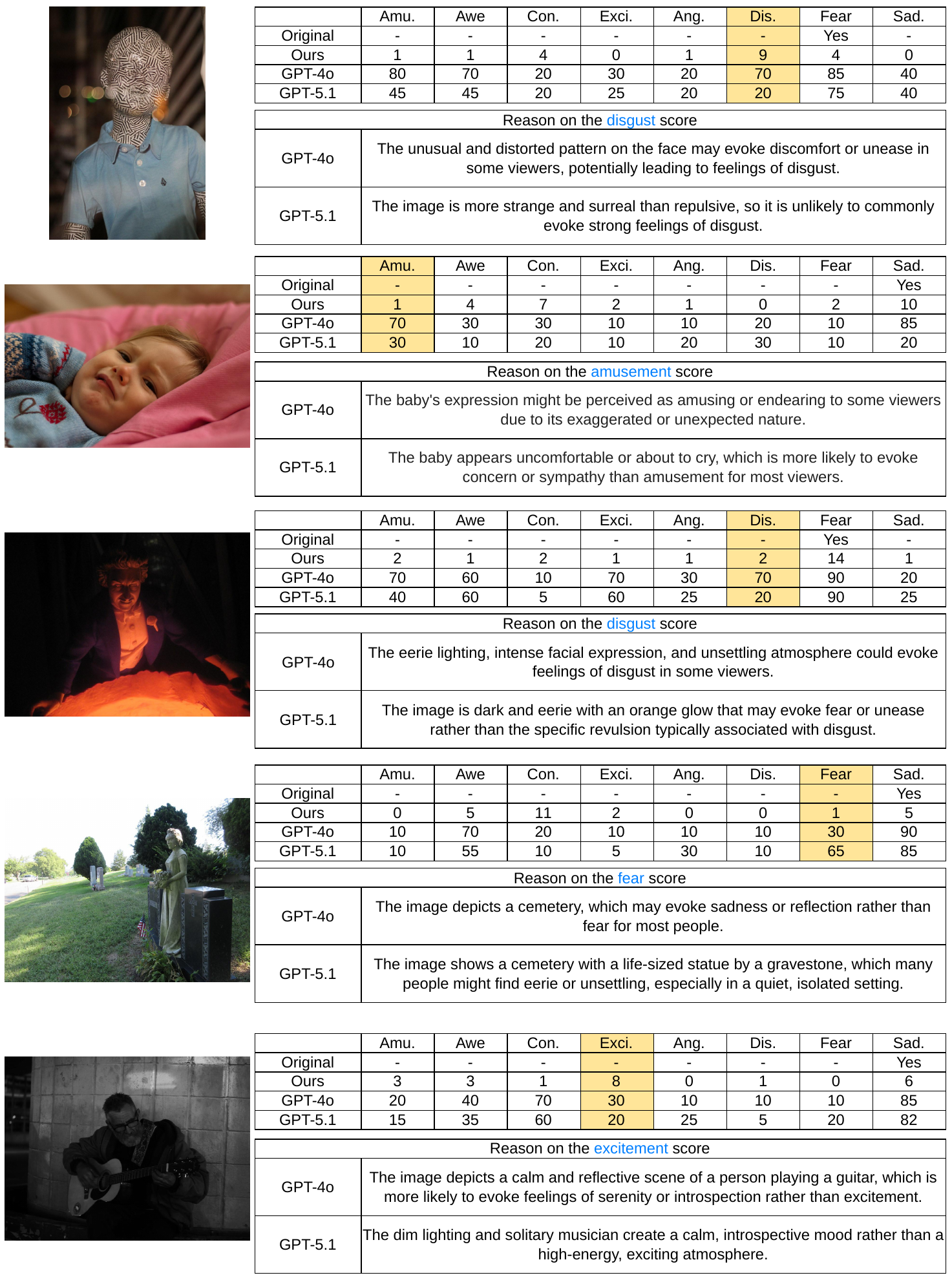}
  \caption{Additional examples of the emotion distribution prediction where the output of GPT-4o and GPT-5.1 are different. The scores in orange are the most different among two models and the labels.}
  \label{fig:examples_dist}
\end{figure}

\begin{figure}[tb]
  \centering
  \includegraphics[width=12cm]{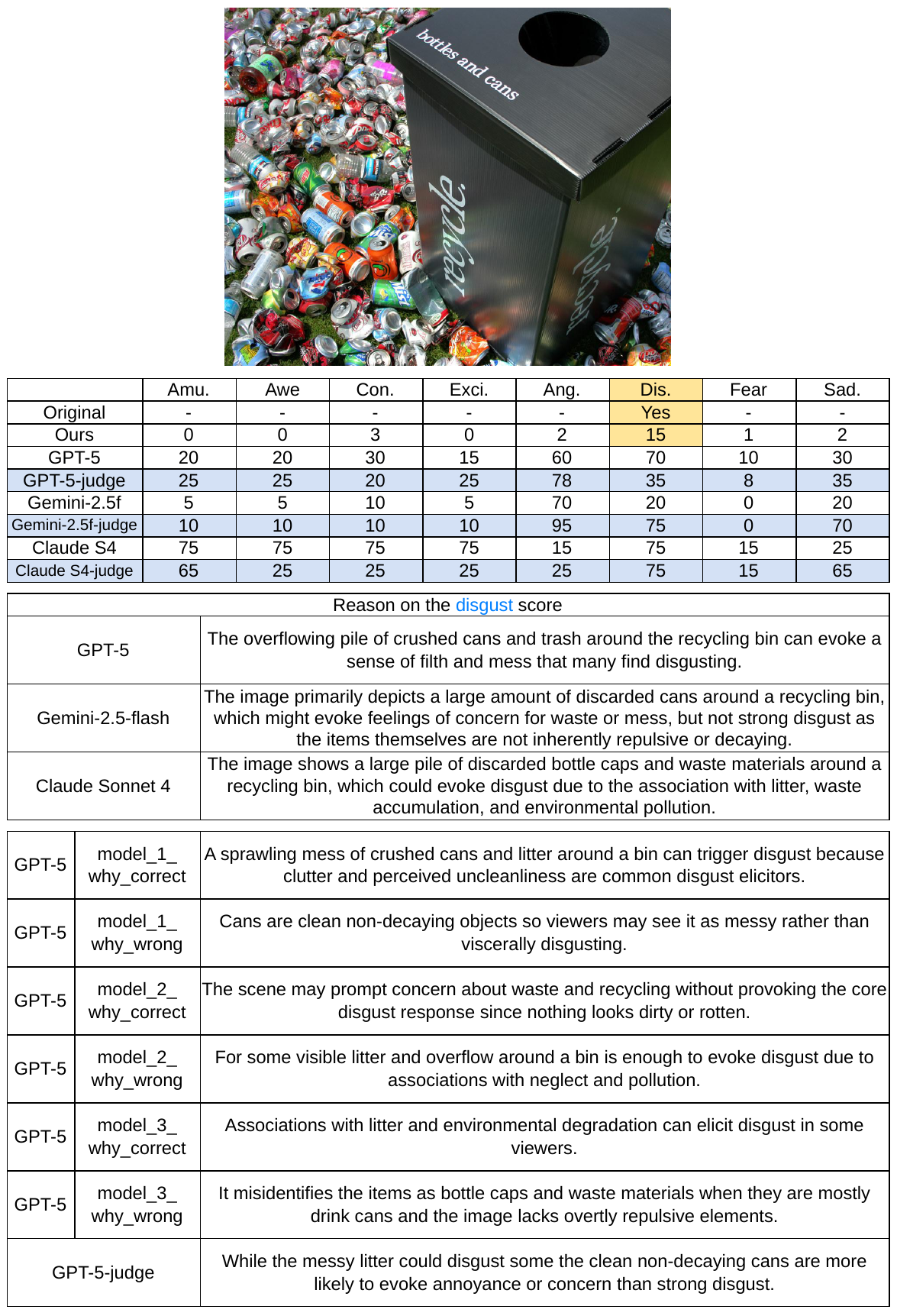}
  \caption{Additional example of the LLM-as-a-judge verification. The GPT-5's prediction on \textit{disgust} is reversed due to the effect of Gemini-2.5-flash, resulting a decreasing performance on emotion distribution prediction.}
  \label{fig:examples_laaj_0}
\end{figure}

\begin{figure}[tb]
  \centering
  \includegraphics[width=12cm]{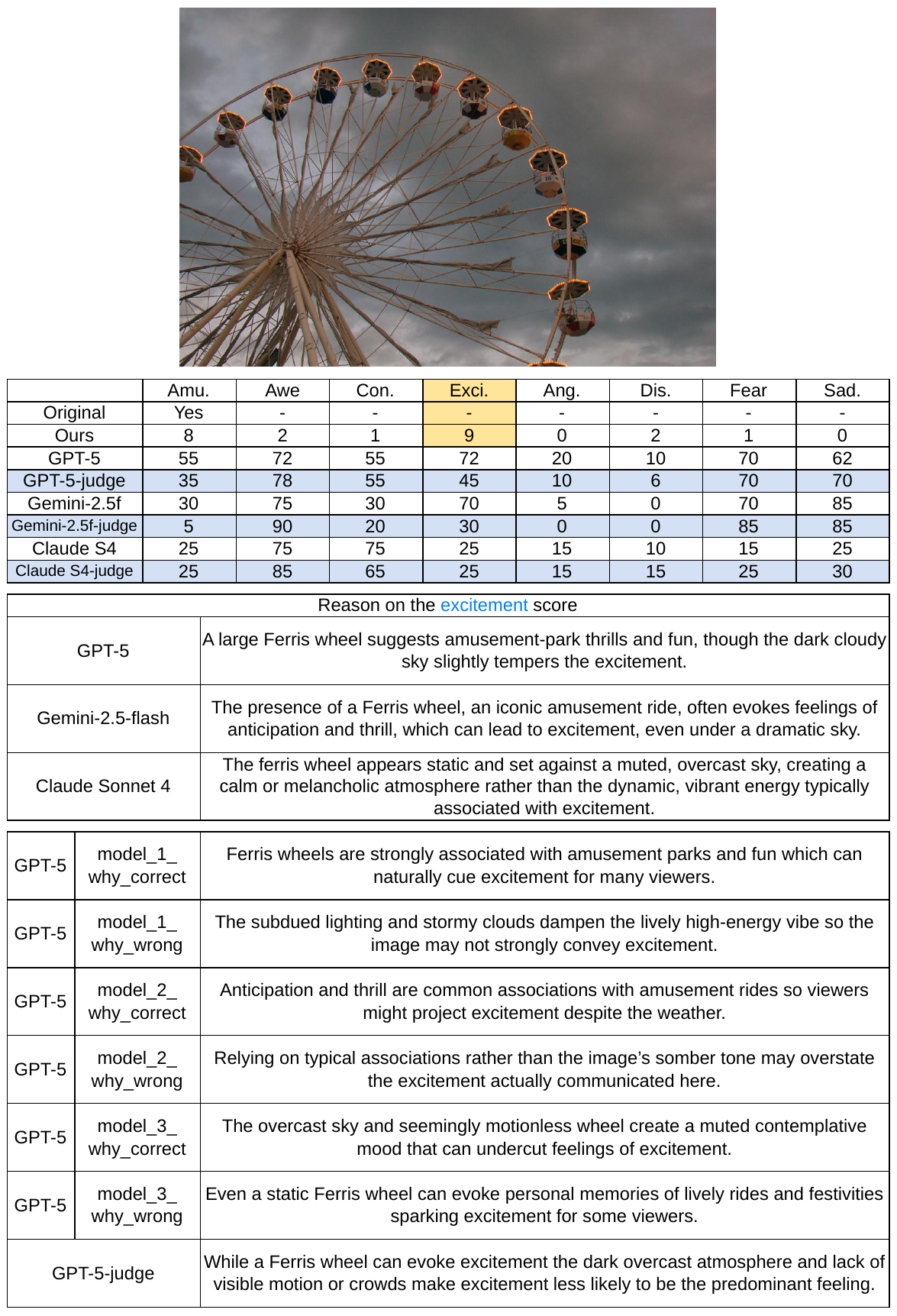}
  \caption{Extra example of the LLM-as-a-judge verification}
  \label{fig:examples_laaj_1}
\end{figure}

\begin{figure}[tb]
  \centering
  \includegraphics[width=12cm]{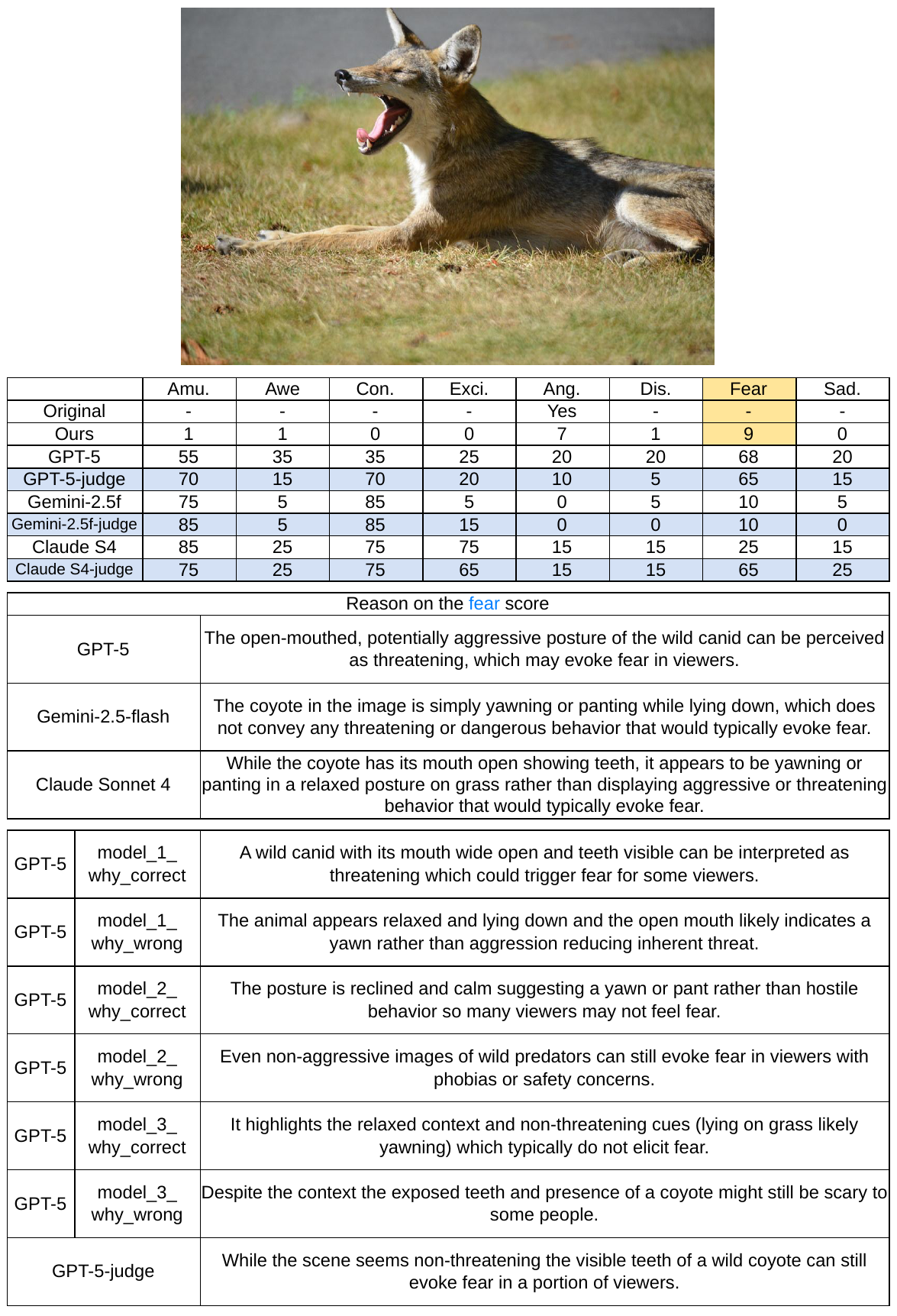}
  \caption{Extra example of the LLM-as-a-judge verification}
  \label{fig:examples_laaj_2}
\end{figure}

We provide additional examples of our annotation and evaluation results in this section. Specifically, \cref{fig:examples_dominant} presents examples of dominant emotion prediction, \cref{fig:examples_dist} presents examples of emotion distribution prediction, and \cref{fig:examples_laaj_0,fig:examples_laaj_1,fig:examples_laaj_2} present examples of the LLM-as-a-judge verification.

\end{document}